\documentclass[10pt,twocolumn,letterpaper]{article}

\usepackage[pagenumbers]{cvpr} 

\usepackage[dvipsnames]{xcolor}

\definecolor{cvprblue}{rgb}{0.21,0.49,0.74}
\usepackage[pagebackref,breaklinks,colorlinks,citecolor=cvprblue]{hyperref}

\def\paperID{4} 
\def\confName{FGVC}
\def\confYear{2024}

\title{Comparing Importance Sampling Based Methods for Mitigating the Effect of Class Imbalance}

\author{Indu Panigrahi\thanks{Equal contribution.} \ and Richard Zhu\footnotemark[1] \\
Princeton University\\
{\tt\small \{indup, ryzhu\}@princeton.edu}
}

\begin{document}
\maketitle
\begin{abstract}

Most state-of-the-art computer vision models heavily depend on data.
    However, many datasets exhibit extreme class imbalance which has been shown to negatively impact model performance.
    Among the training-time and data-generation solutions that have been explored, one subset that leverages existing data is \textit{importance sampling}.
    A significant portion of this work focuses primarily on the CIFAR-10 and CIFAR-100 datasets which fail to be representative of the scale, composition, and complexity of current state-of-the-art datasets.
    In this work, we explore and compare three techniques that derive from importance sampling: loss reweighting, undersampling, and oversampling.
    Specifically, we compare the effect of these techniques on the performance of two encoders on an impactful satellite imagery dataset, Planet's Amazon Rainforest dataset, due to its challenging nature and existing class imbalance.
    Furthermore, we perform supplemental experimentation on a scene classification dataset, ADE20K, to test on a contrasting domain and clarify our results.
    Across both types of encoders, we find that up-weighting the loss for and undersampling has a negigible effect on the performance on underrepresented classes.
    Additionally, our results suggest oversampling generally improves performance for the same underrepresented classes.
    Interestingly, our findings also indicate that there may exist some redundancy in data in the Planet dataset.
    Our work aims to provide a foundation for further work on the Planet dataset and similar domain-specific datasets.
    We open-source our code for future work on other satellite imagery datasets as well.

\end{abstract}    
\section{Introduction}
\label{sec:intro}

Detecting illegal slash-and-burn and mining activities in protected rainforests is crucial for ensuring the well-being of such areas, and satellite imagery can facilitate this process.
Satellite imagery can cover vast areas, so there can exist a large the amount of data to parse.
Classification models can expedite this process by automating the categorization of satellite images according to their captured patterns.

\begin{figure}[h!]
\centering
     \includegraphics[width=\linewidth]{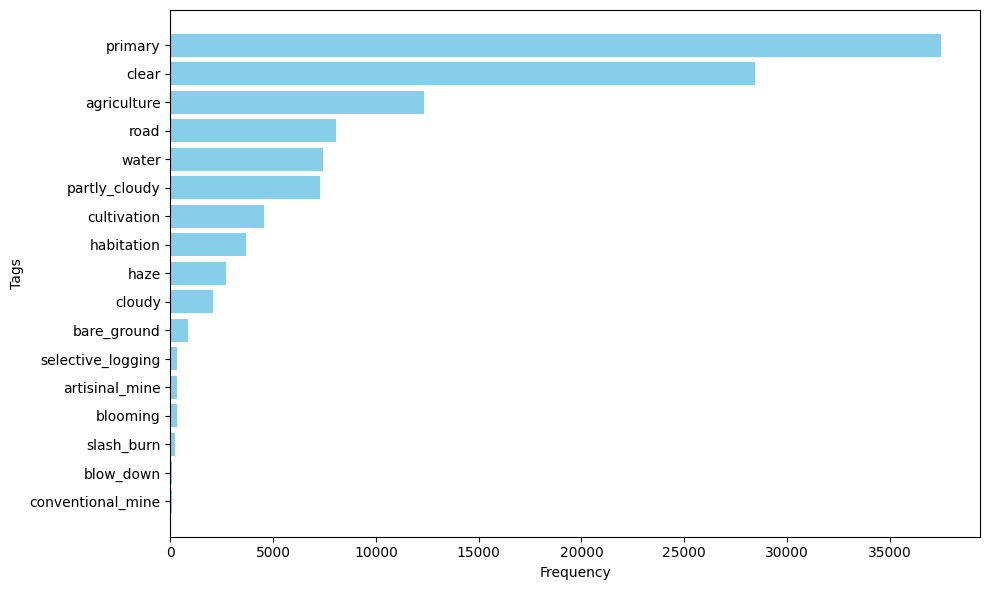}
     \caption{\textbf{Long-tail distribution of classes in the Planet dataset.} The frequency of images across the $17$ labels in the Planet dataset form a skewed distribution where some labels, such as ``primary'' and ``agriculture'', have many images and other labels, such as ``blooming'' and ``blow\_down'', have very few images. Important labels, such as ``slash\_and\_burn'' and ``artisanal\_mining'', are some of the most underrepresented classes in the dataset.}
     \label{fig:for_freq}
\end{figure}

However, many large and widely-used datasets exhibit a long-tail distributions where some classes have significantly more examples than others (Figure \ref{fig:for_freq}) \cite{places365,longtail2,lvis,longtail3,zhou_scene_2017,inaturalist}.
Oftentimes, machine learning models achieve higher accuracies on overrepresented classes than on underrepresented classes, similar to how a human would be better at classifying images corresponding to classes for which they have seen more examples \cite{classimbalance1,classimbalance2,classimbalance3,classimbalance4}.
However, this induced bias can lead to dangerous and unfair model predictions \cite{bias1,bias2,bias3,wang2023pretrainbias}.

We explore a satellite imagery dataset, the Planet Rainforest dataset \cite{labs_planet_2017,rom_planets_dataset_2019}, that demonstrates this trend.
In fact, the detrimental activities that users would want to detect are some of the worst-represented classes in the dataset (Figure \ref{fig:for_freq}).
Additionally, this type of distribution becomes more difficult to remediate when each image has multiple labels, as is the case with the Planet dataset.


From a theoretical machine learning perspective, in addition to bounding generalization error as we evaluate models on test sets, we may also be interested in developing empirical methods for minimizing expected error. 
By understanding the causes of poor generalization in practice, we can gain intuition in order to further constrain bounds when developing models in practice. 

A natural solution to this issue of class imbalance would be to simply collect or generate more data.
Depending on the application, collecting data is often time-consuming and costly, and generating data may not produce sufficiently-realistic representations.
In general, exploring ways to leverage existing data would be beneficial.

Importance sampling covers a subset of methods that recycles existing data to mitigate the effects of class imbalance  \cite{byrd2019,importance1,importance2, importance3,importance4}.
To that end, we explore and compare three importance sampling based methods: loss reweighting, undersampling, and oversampling.

Specifically, we experiment with two datasets from different domains, the Planet Rainforest \cite{labs_planet_2017,rom_planets_dataset_2019} and the ADE20K \cite{zhou_scene_2017} datasets, and with two image encoders, CLIP \cite{clip} and ResNet-18 \cite{resnet}.
We primarily focus on the Planet dataset since we believe that it provides varied, challenging tasks that can accentuate the effects of class imbalance.
However, we include supplemental experimentation on ADE20K, for completeness and to observe how generalizable our findings are. We strive to evaluate our findings on datasets that test a skillset representative of broader model capabilities.  






Across the two types of encoders, we find that up-weighting the loss on underrepresented classes has a negligible effect on performance.
Additionally, we find that oversampling appears to be more effective than undersampling.
That being said, our results suggest that undersampling could still be a valuable method if we mindfully select samples rather than at random.
Lastly, our experiments indicate that there may exist some redundancy in the Planet dataset which motivates new avenues for future work.
\section{Related Work}
\label{sec:related}

\paragraph{Class Imbalance Problem.}
Class imbalance occurs when some classes have significantly more examples than others in a dataset. 
Imbalanced distributions naturally arise across many types of datasets ranging from medical imaging to environmental satellite imagery \cite{kubat_machine_1998,rao_data_2006,wei_effective_2013,herland_big_2018}.
While other forms of imbalances exist (e.g., object-level for object detection and pixel-level for segmentation), our work explores solutions for class imbalance since we examine image classification. 
While the class imbalance problem has been well-studied in traditional non-deep machine learning and computer vision, recent studies have noted the lack of work on studying the effects of and methods for counteracting class imbalance across a variety of datasets and imbalance levels \cite{classimbalance2}.
We work with a satellite imagery and a scene classification dataset to evaluate on different domains.
Prior work has shown class imbalance to deteriorate model performance on underrepresented classes \cite{thai-nghe_improving_2009,luque_impact_2019,gong_rhsboost_2017,prati_class_2015}.
However, these classes can be extremely important, as is the case with the Planet dataset (Figure \ref{fig:for_freq}).


\paragraph{Data generation.}
A natural solution to the class imbalance problem would be to collect more data to balance the dataset.
Since manual data collection can be quite arduous for specialized applications (e.g., medicine, botany, and geosciences) \cite{botany,geosciences,medicine,inaturalist}, an alternative is to generate additional data via self-supervised learning \cite{selfsuplearning,selfsuplearning2}, semi-supervised learning \cite{lee2013,semisuplearning}, or generative adversarial networks (GANs) \cite{gans1,gans2}.
However, one common limitation is that the generated data can sometimes be low in quality, particularly for specialized applications such as for those represented in this paper (e.g. satellite imagery of rainforests).
Furthermore, leveraging existing data would still be data-efficient but simpler.
As a result, we choose to focus on comparing methods that leverage existing data.

\paragraph{Cost-sensitive learning.}
One such method is cost-sensitive learning.
Cost-sensitive learning is a progressive training method where higher-cost examples contribute more significantly in updating the model. In practice, this means the user can tailor the model to heavily penalize certain misclassifications \cite{fernandez_cost-sensitive_2018, zadrozny_cost-sensitive_2003}. 
For example, if the model performs poorly on an underrepresented class, the user can increase the penalty for mistakes on that class by influencing the cost (i.e., loss).
In this work, we implement a loss reweighting scheme which we describe in Section \ref{sec:methods}.


\paragraph{Importance Sampling.}
Another related subset of methods that overlaps with cost-sensitive learning is importance sampling.
Importance sampling is useful when the test distribution $q$ is not accurately reflected in the training distribution $p$ \cite{importance1,importance2,importance3,importance4}.
For some example $x$, if we are estimating some function $f(x)$, importance sampling weights $x$ based on its likelihood ratio $\frac{q(x)}{p(x)}$ as shown in Equation \ref{eq1} \cite{byrd2019}.
\begin{align}
\label{eq1}
    \mathbb{E}_p \left[ \frac{q(x)}{p(x)} f(x) \right] &= \int_x \frac{q(x)}{p(x)} f(x) p(x) dx \\
    &= \int_x q(x) f(x) dx \\
    &= \mathbb{E}_q [f(x)]
\end{align}
The idea of importance sampling forms the basis for the three methods that we explore which we describe in the next section.
\section{Methods}
\label{sec:methods}
Our work explores and compares loss reweighting and two types of resampling: undersampling and oversampling.
In this section, we provide a description of each method.

\begin{figure}[h!]
\centering
        \centering
         \includegraphics[width=0.6\linewidth]{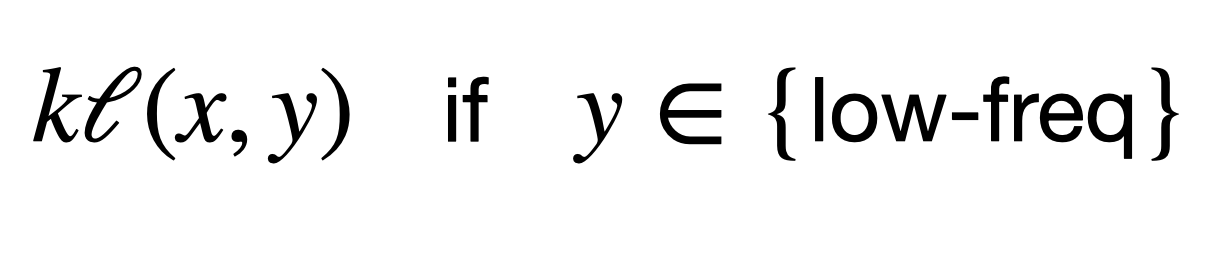}
         \caption{\textbf{Reweighting Loss Method.} Our approach scales the loss $\ell(x,y)$ where $x$ is the image and $y$ is the label by a constant factor $k$ if the instance belongs to a low-frequency class. We evaluate with a range of values for $k$ (Section \ref{sec:results}), specifically $1.5$, $2$, $2.5$, $3$, $5$, and $10$.}
         \label{fig:reweight_method}
\end{figure}

\paragraph{Reweighting loss.} 
This technique reweights the loss at training time to amplify the loss for low-frequency classes and/or diminish the loss for other classes.
One approach scales the loss of all low-frequency classes by a constant (Figure \ref{fig:reweight_method}), while another scales the loss of all classes relative to their frequency. 
We choose to do the former because we find that it offers more granularity in controlling which classes are amplified in situations where two labels may be very similar. 
For example, if we have an image classification task where the classes "roadway" and "highway" appear moderately frequently, we may still wish to consider these classes as low-frequency classes to improve the model's ability to distinguish between the two (likely very similar) images.

Prior work has previously implemented variants of the latter approach.
One such work evaluates the effectiveness of learning a mapping from classes to reweighting factors \cite{shu_meta-weight-net_2019}.
Using an MLP with a single hidden layer, they observe some performance gains on the CIFAR-10 and -100 datasets, achieving test accuracies on ResNet-32 that are greater than or equal to those on a class-balanced version of the dataset. 
Another work simply scales the loss by the inverse class frequency.
Their method, \texttt{class-balanced focal loss}, achieves a nominally better training loss on the \texttt{ILSVRC 2012} \cite{imagenet} and \texttt{iNaturalist 2018} \cite{inaturalist} datasets after sufficiently many epochs, while results at earlier epochs underperform the default softmax cross-entropy \cite{cui_class-balanced_2019}. 
On CIFAR-10 and -100, they report gains, albeit marginal ones.
We include a discussion about another related work that takes the latter approach in the context of our results for completeness (Section \ref{sec:results}) \cite{byrd2019}.

\begin{figure}
    \centering
    \begin{subfigure}[b]{0.9\linewidth}
         \centering
         \includegraphics[width=\linewidth]{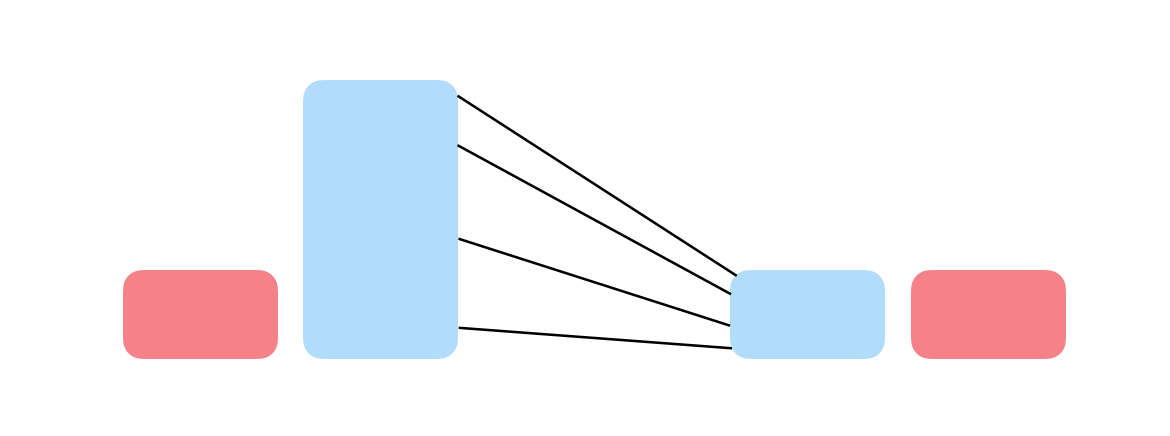}
         \caption{\textbf{Undersampling.}}
         \label{fig:undersample}
    \end{subfigure}
    
    \begin{subfigure}[b]{0.9\linewidth}
         \centering
         \includegraphics[width=\linewidth]{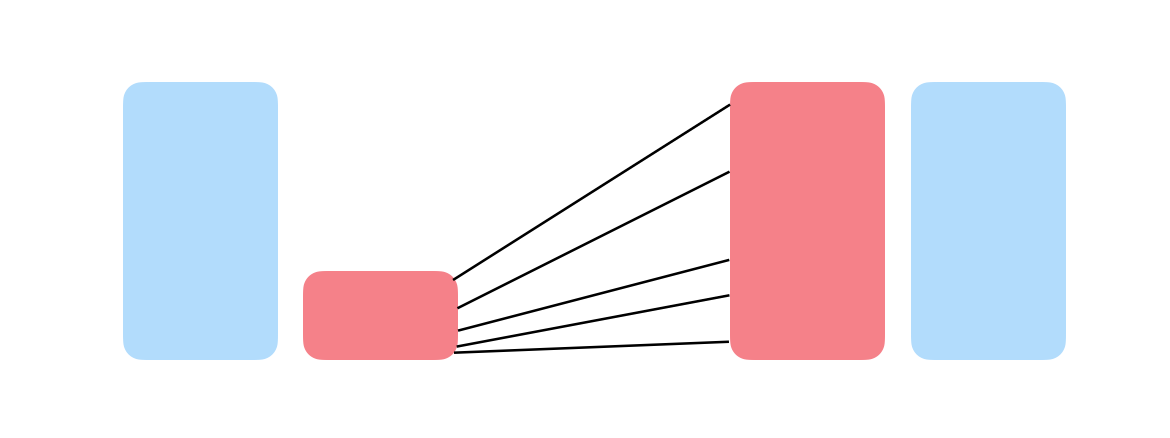}
         \caption{\textbf{Oversampling.}}
         \label{fig:oversample}
    \end{subfigure}
    \caption{\textbf{Types of Resampling.} Oversampling (b) trains on repeated elements from underrepresented classes, while undersampling (a) does the opposite, training on a randomly-selected subset of elements from overrepresented classes. The threshold chosen for the number of images to resample or remove is dataset-dependent.}
    \label{fig:resample}
\end{figure}

\paragraph{Undersampling.} 
Undersampling reduces the number of examples in overrepresented classes by randomly sampling a subset of a predefined size (Figure \ref{fig:undersample}).
This size is usually the size of the lowest-frequency class \cite{classimbalance1,undersamplethreshold}.
This method effectively decreases the total number of elements.
As a result, the number of examples seen from each class during training is generally more uniform, so intuitively, the model is expected to perform more uniformly across classes.
One of the first papers that discussed class imbalance highlighted the potential performance improvements of undersampling methods \cite{kubat_addressing_2000}.

Since undersampling removes examples from the training set, this method appears to be more suitable for large datasets where we can afford to prune the dataset without significantly affecting results \cite{classimbalance1}.
While the exact threshold for such a decision seems nebulous, most prior work finds that the thresholds that we use (typically undersampling all elements to a threshold approximately equal to the number of elements in the least populous class being trained on), does not significantly decrease performance \cite{classimbalance1}.
Thus, applying undersampling to the Planet dataset seems reasonable.

\paragraph{Oversampling.} 
Similar to undersampling, oversampling aims to create a more uniform distribution of training examples across classes.
However, this process instead repeats training examples from low-frequency classes via random sampling until those classes reach a comparable frequency to the other classes (Figure \ref{fig:oversample}).
Both undersampling and oversampling can be tricky for multi-label classification (e.g. with the Planet dataset).
There exist several approaches to ensure a roughly similar number of elements in both resampling methods, and we discuss our approach in the next section (Section \ref{sec:experiments}).

\section{Experimental Details}
\label{sec:experiments}



\begin{figure}[h!]
\centering
        \centering
         \includegraphics[width=\linewidth]{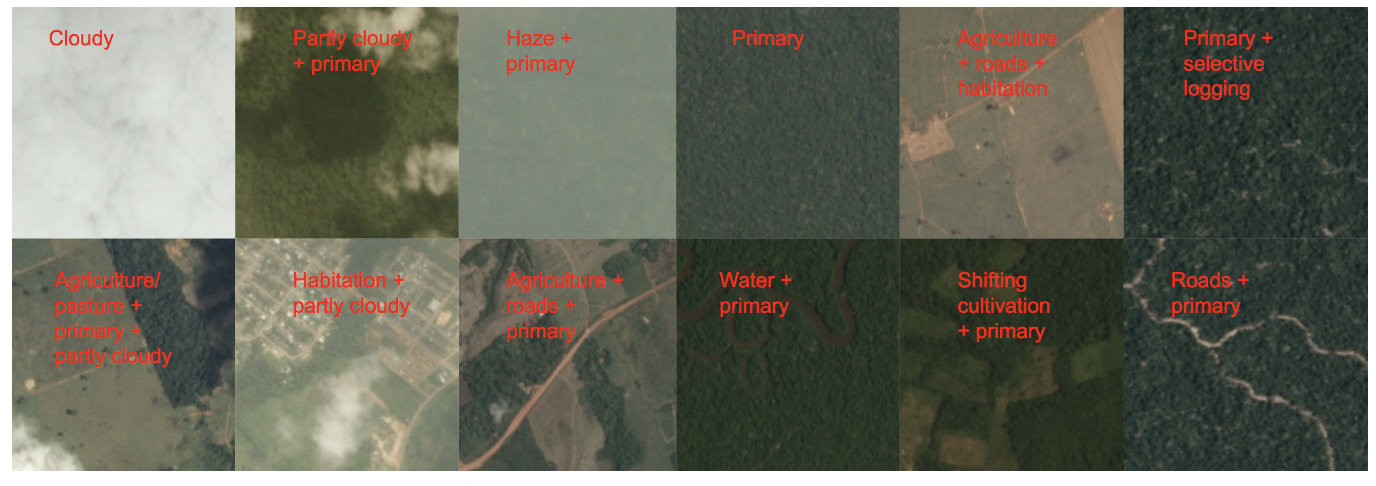}
         \caption{\textbf{Sample Images from the Planet dataset.} Figure from Planet's Kaggle website \cite{labs_planet_2017}. The Planet dataset consists of fine-grain categories such as ``conventional mining'' and ``artisanal mining''. Each image has multiple labels one of which is a cloud cover label, such as ``partly cloudy'' or ``clear''.}
         \label{fig:planet_sample}
\end{figure}

\begin{figure}[h!]
        \centering
         \includegraphics[width=\linewidth]{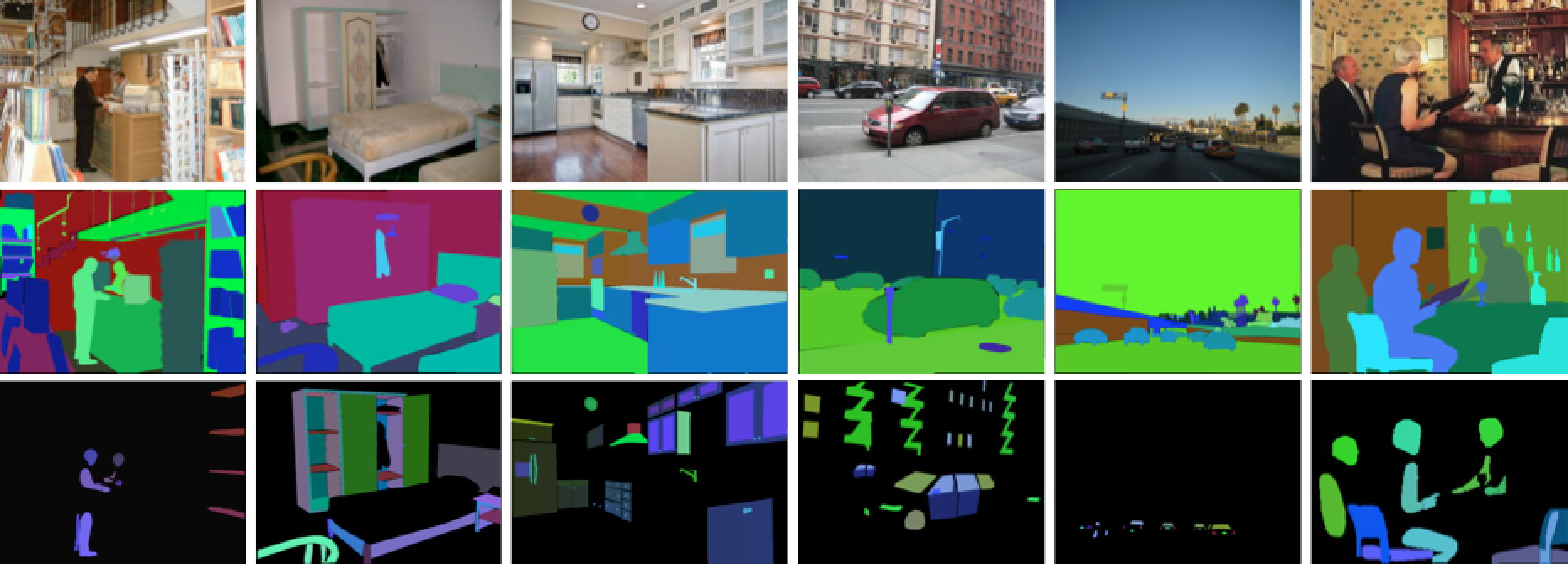}
         \caption{\textbf{Sample Images from the ADE20K dataset.} The ADE20K dataset is a scene classification dataset with one label per image. ADE20K includes object-level and segmentation labels as shown; however, we only use the image classification labels. Figure from the ADE20K website \cite{zhou_scene_2017}.}
         \label{fig:ade20k_sample}
\end{figure}

\paragraph{Data and Tasks.}
We primarily experiment with the Planet Rainforest dataset (Figure \ref{fig:planet_sample}) and include supplemental experimentation on the ADE20K dataset (Figure \ref{fig:ade20k_sample}).
The Planet dataset contains satellite imagery of the Amazon Basin in South America, and each image has 4 bands (red, green, blue, and near-infrared).
This dataset consists of fine-grain categories such as ``conventional mining'' and ``artisanal mining''.
Each image includes a label that indicates cloud coverage (e.g., ``clear'', ``partly\_cloudy'') along with labels that describe the areas in the images (Figure \ref{fig:planet_sample}).
As a result, each image in the Planet dataset typically has at least two labels.

\begin{figure}[h!]
     \centering
     \includegraphics[width=\linewidth]{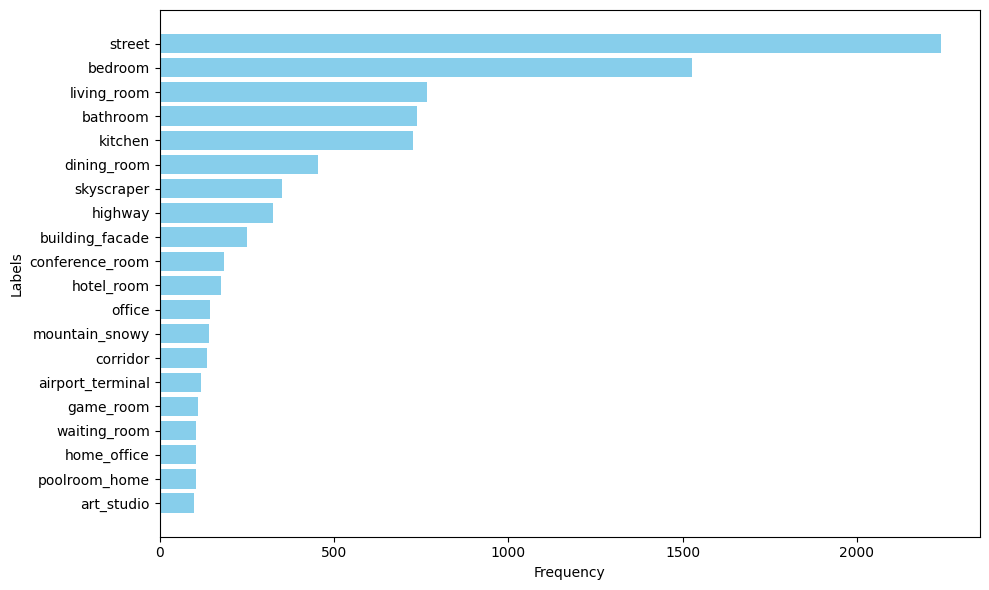}
     \caption{\textbf{Distribution of classes in ADE20K.}}
     \label{fig:ade_freq}
\end{figure}

This dataset demonstrates a long-tail distribution, and the classes that indicate illegal or dangerous forest activites (e.g., ``slash\_and\_burn'' and ``artisanal mining'' are among the lowest-frequency classes (Figure \ref{fig:for_freq}).
This distribution is problematic because machine learning models more generally tend to perform poorly on underrepresented classes.

\vfill
\vfill

The ADE20K dataset contains images of commonplace scenes and has a few levels of annotations.
There are scene classification labels where each image maps to one label and segmentation annotations for the individual objects in each image.
In this work, we focus on image classification, so we use only the scene classification labels.
The ADE20K dataset also demonstrates a long-tail distribution.


We choose the Planet dataset because of the challenging skillset involved and class imbalance, consequently testing multi-label classification.
We include ADE20K in our reweighting experiments to disentangle the results by testing single-label classification (Section \ref{sec:results}) and provide a contrasting domain.
However, the primary focus of this work is on the Planet dataset.


We use all $17$ classes from Planet and select the $20$ most frequent classes (out of the $150$ total) from ADE20K.
Both datasets exhibit long-tail distributions with Planet having the more severe of the two (Figure \ref{fig:for_freq}, \ref{fig:ade_freq}).
Since these are challenge datasets, the test sets are not publicly available, so we apply a $60$-$20$-$20\%$ split to the training and validation sets combined to produce our training, validation, and test sets.
In our results, we report the test accuracies.

\paragraph{Define low-frequency classes.}
Prior works mostly tailor the definition of low-frequency classes to their datasets by selecting classes based on certain features or by predefining a frequency threshold \cite{choosingrare,choosingrare2}.
A more general option is to define the low frequency classes to be those that fall one standard deviation below the mean.
However, for Planet, the mean frequency is $5464.82$, and the standard deviation is $8214.73$ which is too large to follow this standard.
Instead, we define the low-frequency classes to be those that have frequencies that fall within the $25^{th}$ percentile of all the frequencies.
To maintain consistency, we use the same definition to select the low-frequency classes in ADE20K.

\paragraph{Resampling.}
For undersampling, we choose a threshold of $200$ for Planet because this is approximately the frequency of the lower-frequency classes.
Thus, we randomly sample from classes with frequencies that exceed this threshold to reduce those classes to the threshold.

For oversampling, we choose a threshold of $30,000$ for Planet because this is approximately the frequency of the higher-frequency classes.
Similar to undersampling, we randomly sample from classes with frequencies lower than this threshold until they reach the threshold.
However, before doing so, we ensure that all the images from those classes are included so that each image appears at least once.

\begin{figure}[h!]
\centering
    \begin{subfigure}{\linewidth}
        \centering
         \includegraphics[width=\linewidth]{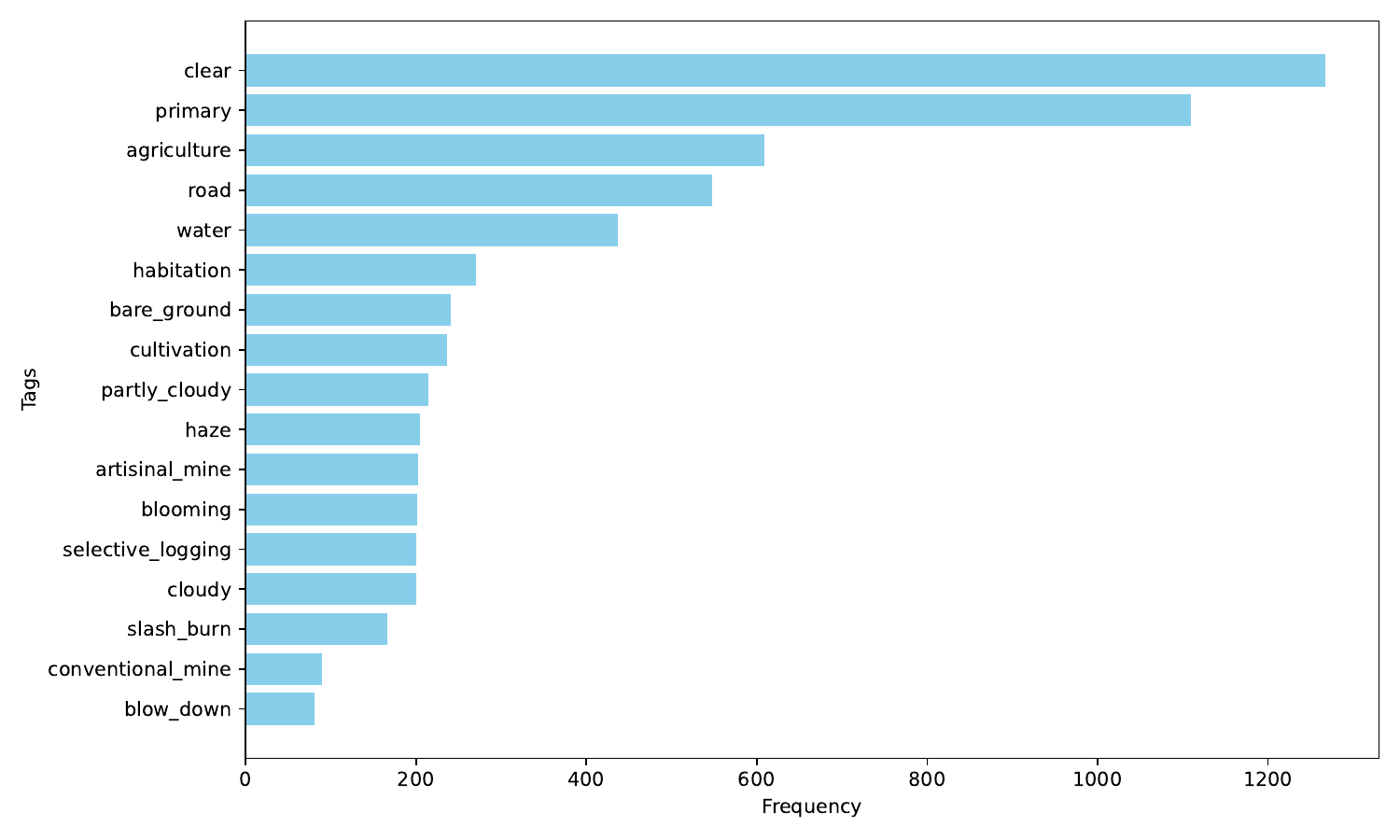}
         \caption{\textbf{Undersampled distribution of the Planet dataset.}}
         \label{fig:for_undersamp_dist}
    \end{subfigure}
    \vspace{\baselineskip}
    \vfill
    \begin{subfigure}{\linewidth}
        \centering
         \includegraphics[width=\linewidth]{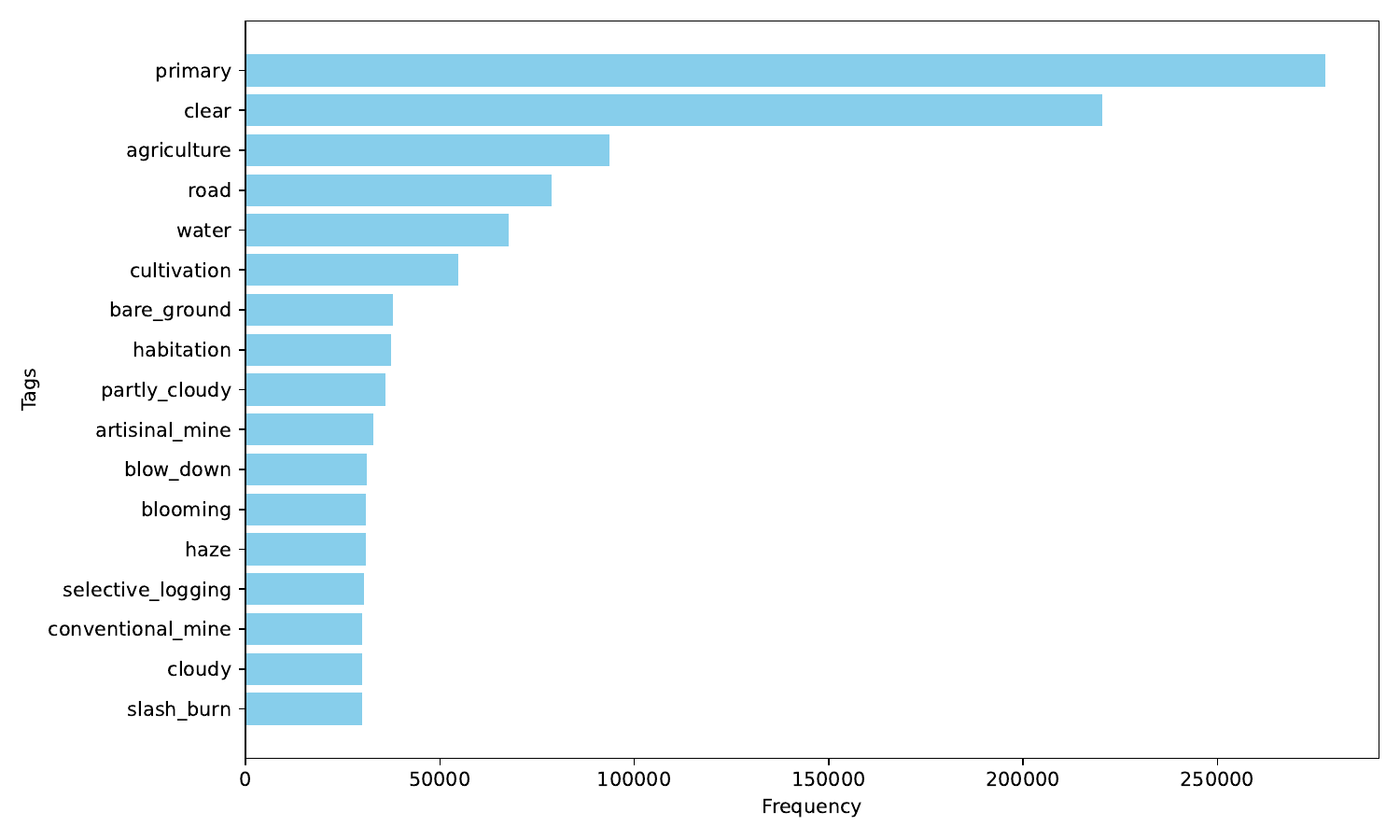}
         \caption{\textbf{Oversampled distribution of the Planet dataset.}}
         \label{fig:for_oversamp_dist}
    \end{subfigure}

    \caption{\textbf{Resampled distributions of the Planet dataset.}}
    \label{fig:for_resample_dist}
\end{figure}

Since the images in Planet can have multiple labels, ensuring a perfectly uniform distribution is unreasonable.
In particular, the "primary" label indicates the presence of trees, and "clear" indicates that there exists no cloud cover.
As a result, these two labels overlap with many images, especially as each image has a cloud cover label (i.e., "clear", "partly\_cloudy", "haze").
However, we achieve a generally uniform distribution among the more disjoint classes (Figure \ref{fig:for_resample_dist}) which is significantly less skewed than the original distribution (Figure \ref{fig:for_freq}).

\begin{figure}[h!]
        \centering
         \includegraphics[width=\linewidth]{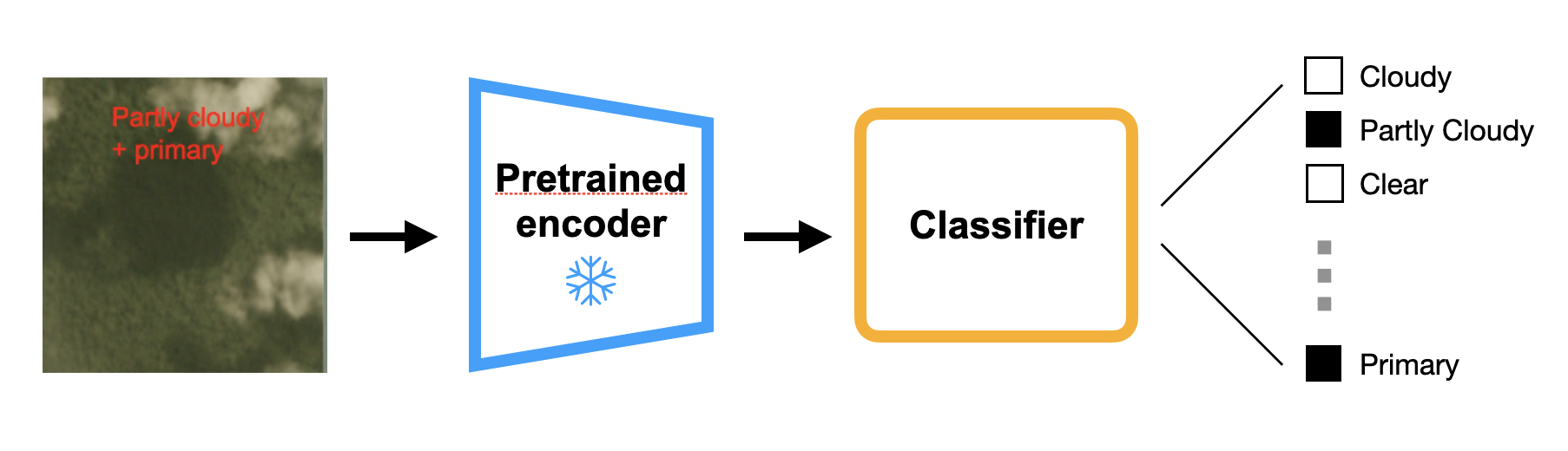}
         \caption{\textbf{Model Architecture.} We implement a simple classifer that takes as input embeddings of the input images generated by a frozen, pretrained encoder. We experiment with CLIP \cite{clip} and ResNet-18 \cite{resnet} as encoders.}
         \label{fig:model}
\end{figure}

\paragraph{Architecture.}
Our architecture consists of a pretrained encoder followed by a simple classifier (Figure \ref{fig:model}).
We experiment with two encoders pretrained on ImageNet: CLIP \cite{clip} and ResNet-18 \cite{resnet}.
We implement a simple classifier in PyTorch that consists of the following layers: Linear, ReLU, Linear, and Sigmoid.
We use a binary cross-entropy loss function.
The encoders are frozen, so we specifically train the classifier.

\paragraph{Training.}
After performing a hyperparameter search, we decide to train each model for $20$ epochs using a learning rate of $1e-3$, batch size of $64$, and dropout rate of $0.2$.
We train all the models with the same hyperparameters to control for variance in performance caused by hyperparameters as do prior works.
\section{Results \& Discussion}
\label{sec:results}

We make 3 key observations. First, that upweighting has a small effect on class performance. While we observe this trend in experimental settings where we exclusively up-weight low-frequency classes, and leave mid- and high-frequency classes unchanged, we believe that a similar trend would exist even if all classes were oversampled to a constant frequency. This conclusion is supported by the largely similar proportional sampling techniques from prior works (see Section \label{sec:methods}) that show only nominal performance gains. Second, we find that undersampling has a similarly small effect on class performance and even (albeit somewhat unsurprisingly since parts of the datasets are being omitted from the model's training data) reduces performance on mid- and high-frequency classes. We lastly observe, a similar trend in oversampling. Most surprisingly, we note that the traditionally well-accepted and intuitive methods of reducing the effect of dataset bias appear to be largely ineffective with current or recent state of the art models.

In order to investigate the generalizability of these results, we consider performance across two different datasets (Planet and ADE20K) which have drastically different features. Their label distributions differ (see Figure \ref{fig:for_resample_dist}), and their contrast, color distribution, and purpose vary (see Figure \ref{fig:planet_sample} and \ref{fig:ade20k_sample}). Though we are primarily interested in dataset characteristics, we also attempt to study the generalizability of our finding with respect to the models themselves. We find our conclusions to hold fairly well between both CLIP and ResNet-18 models.

\subsection{Up-weighting the loss for low-frequency classes does not have a significant effect on the performance on any of the classes.}
\begin{figure}[h!]
\centering
    \begin{subfigure}{\linewidth} 
         \centering
         \includegraphics[width=\linewidth]{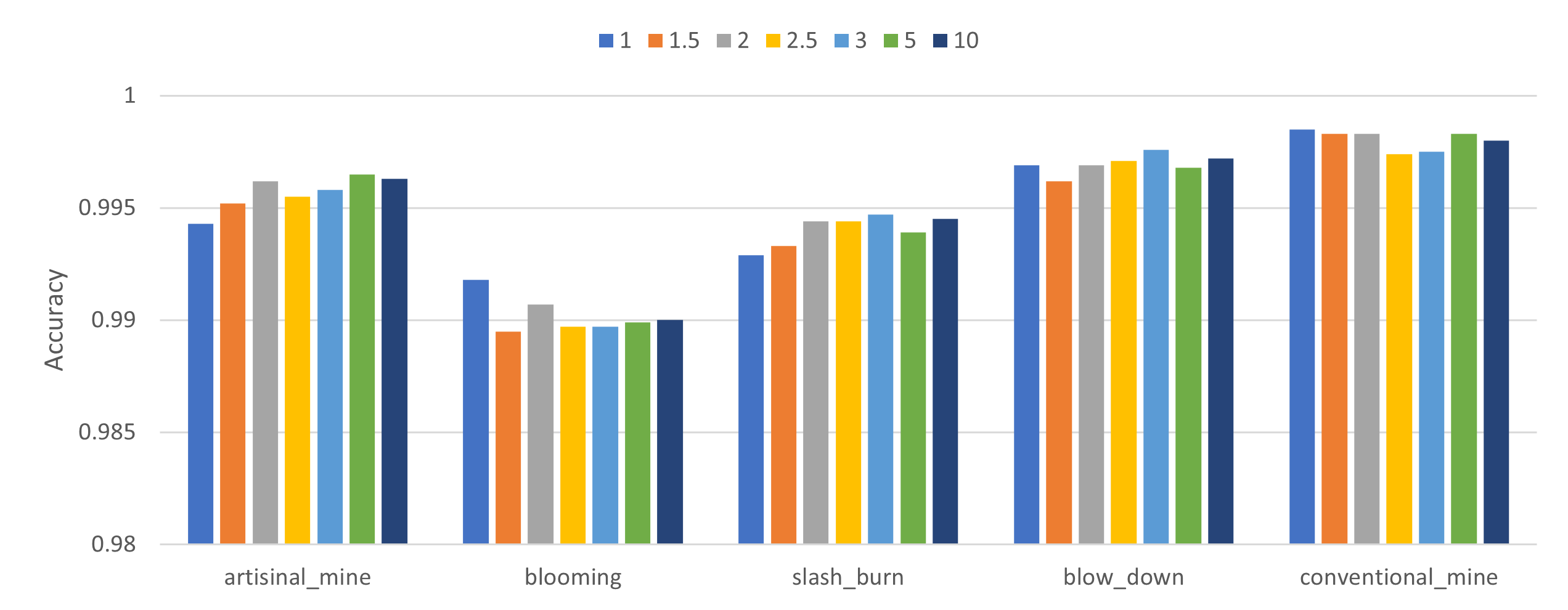}
         \caption{\textbf{Up-weighting accuracy with CLIP encoder.}}
         \label{fig:loss_results_clip}
    \end{subfigure}
    
    \begin{subfigure}{\linewidth}
         \centering
         \includegraphics[width=\linewidth]{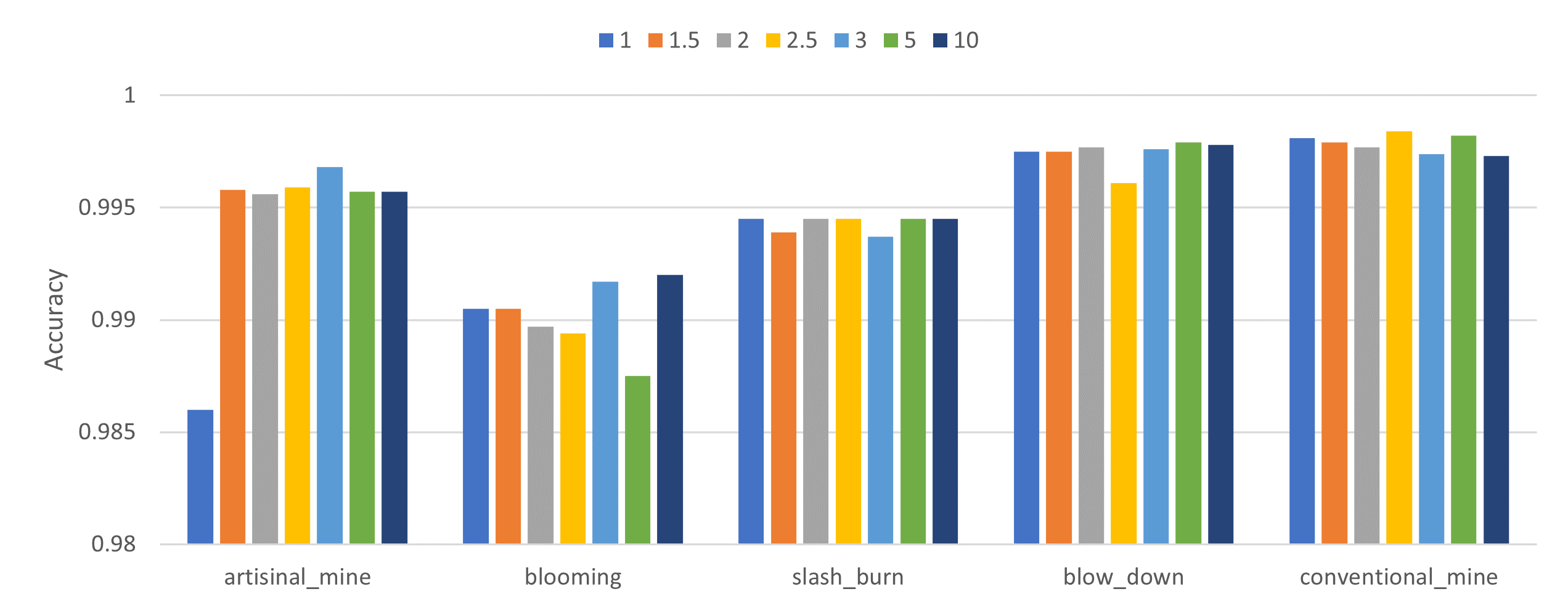}
         \caption{\textbf{Up-weighting accuracy with ResNet-18 encoder.}}
         \label{fig:loss_results_resnet}
    \end{subfigure}
    \caption{\textbf{Up-weighting accuracy on the Planet dataset on low-frequency classes.} These graphs depict the accuracies when scaling the loss for low-frequency classes by factors of $1.5, 2, 2.5, 3, 5, $ and $10$ with different encoders; the factor of $1$ corresponds to the baseline accuracy. Both panels depict the accuracies on the low-frequency classes only; see Figure \ref{fig:loss_results_for_nr} for the remaining classes. For both encoders, we find that there exists no consistent trend across scaling factors or within a factor.}
    \label{fig:loss_results_planet}
\end{figure}

\begin{figure}[h!]
    \centering
    \begin{subfigure}{\linewidth}
    \centering
            \includegraphics[width=\linewidth]{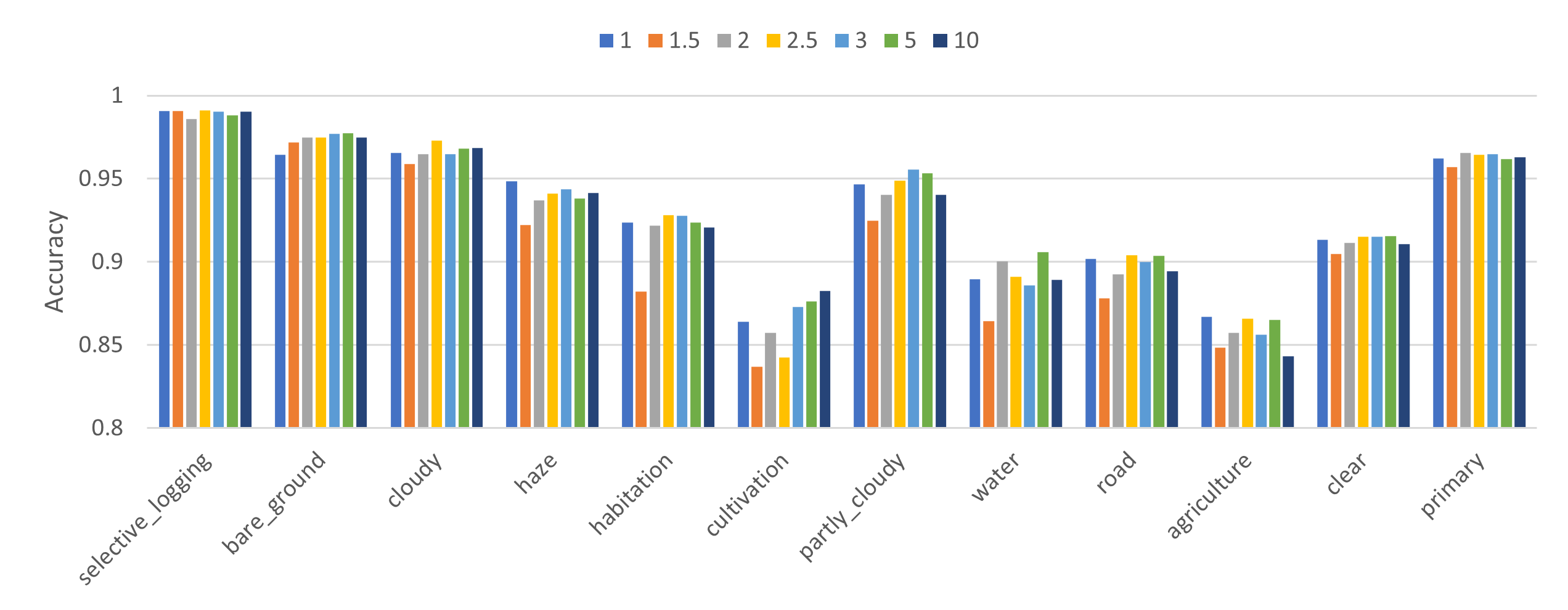}
            \caption{\textbf{Up-weighting accuracy on the remaining classes in Planet when using CLIP.}}
            \label{fig:clip_for_loss_nr}
    \end{subfigure}

    \vfill
    \vspace{\baselineskip}

    \begin{subfigure}{\linewidth}
    \centering
         \includegraphics[width=\linewidth]{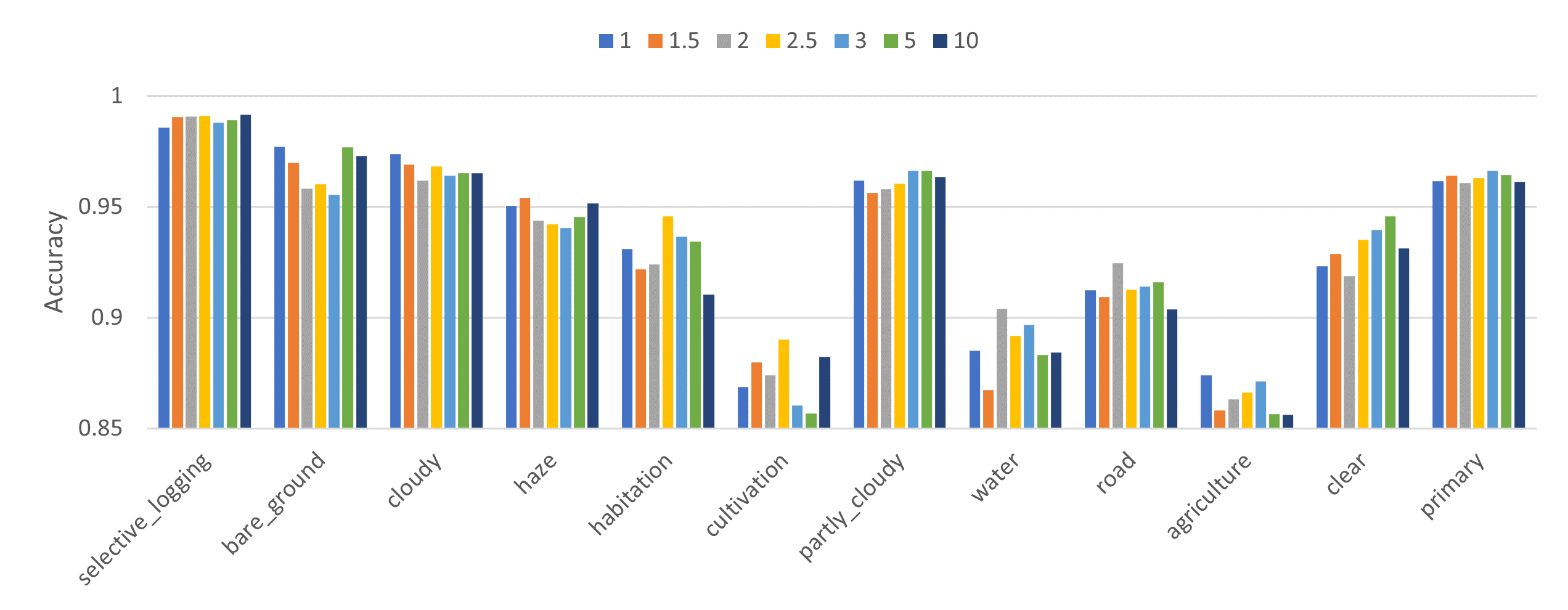}
        \caption{\textbf{Up-weighting accuracy on the remaining classes in Planet when using ResNet-18.}}
         \label{fig:resnet_for_loss_nr}
    \end{subfigure}
    \caption{\textbf{Up-weighting accuracy on Planet on remaining classes.} These graphs depict the accuracies on the remaining classes in Planet when scaling the loss for low-frequency classes by factors of $1.5, 2, 2.5, 3, 5, $ and $10$; the factor of $1$ corresponds to the baseline accuracy. Panel (a) corresponds to using CLIP and panel (b) corresponds to using ResNet-18.}
    \label{fig:loss_results_for_nr}
\end{figure}

\begin{figure}[h!]
    \centering
         \centering
         \includegraphics[width=\linewidth]{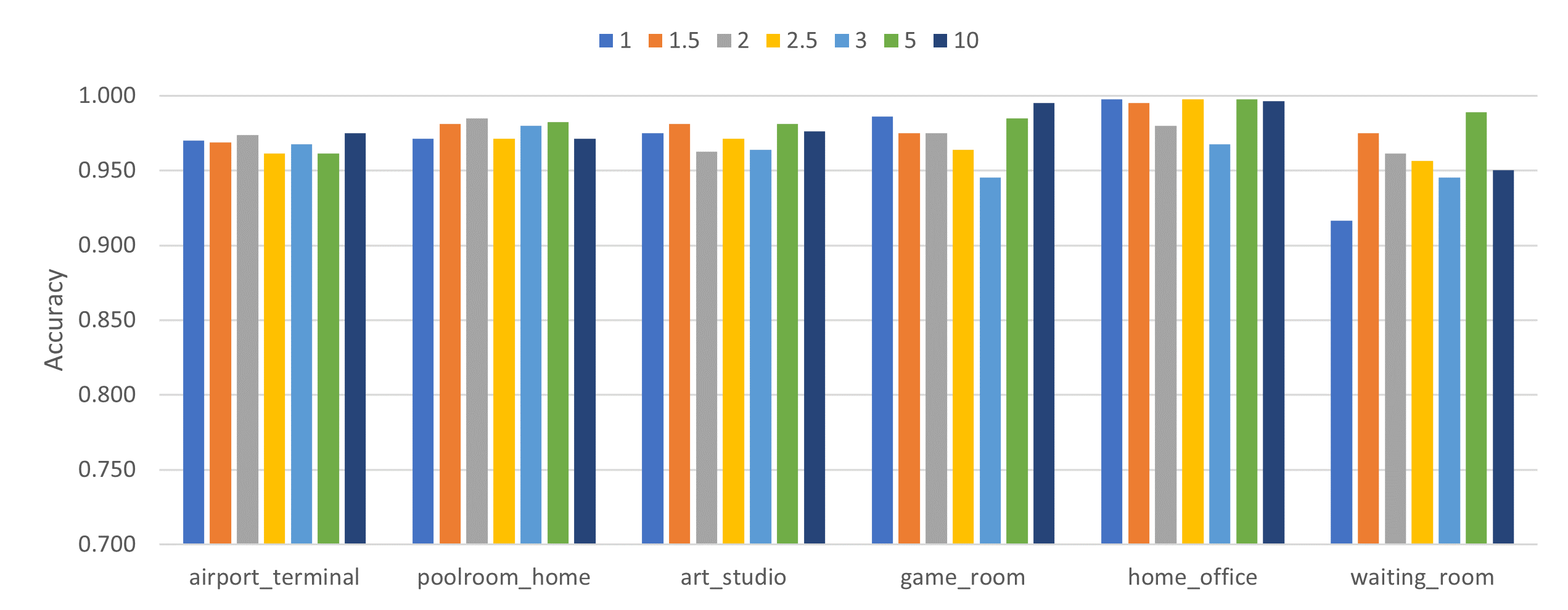}
    \caption{\textbf{Up-weighting accuracy on ADE20K with CLIP encoder on low-frequency classes.} This graph depicts the accuracies when using a CLIP encoder and scaling the loss for low-frequency classes by factors of $1.5, 2, 2.5, 3, 5, $ and $10$. The factor of $1$ corresponds to the baseline accuracy. Similar to the results from the Planet dataset, we find that there is no consistent trend across the scaling factors.}
    \label{fig:loss_results_ade20k}
\end{figure}

Across both models, we find that the performance does not exhibit a consistent improvement or decay when we up-weight the loss of the low-frequency classes in Planet (Figure \ref{fig:loss_results_planet}, \ref{fig:loss_results_for_nr}).
The results for multi-label classification should be interpreted with the caveat that each image can have more than one label.
For example, if an image has the labels "primary" and "slash\_and\_burn", then we would up-weight the loss on the image because "slash\_and\_burn" is a low-frequency class.
However, in doing so, we implicitly up-weight the loss on "primary".
Thus, the up-weighting is not cleanly done on one class and can cause sporadic drops or gains in performance.
To disentangle these effects, we check the performance of the same up-weighting scheme on single-label classification.
However, we again find that the performance of the models does not significantly change when we up-weight the loss of the low-frequency classes (Figure \ref{fig:loss_results_ade20k}).

\begin{figure}[h!]
    \centering
    \includegraphics[width=\linewidth]{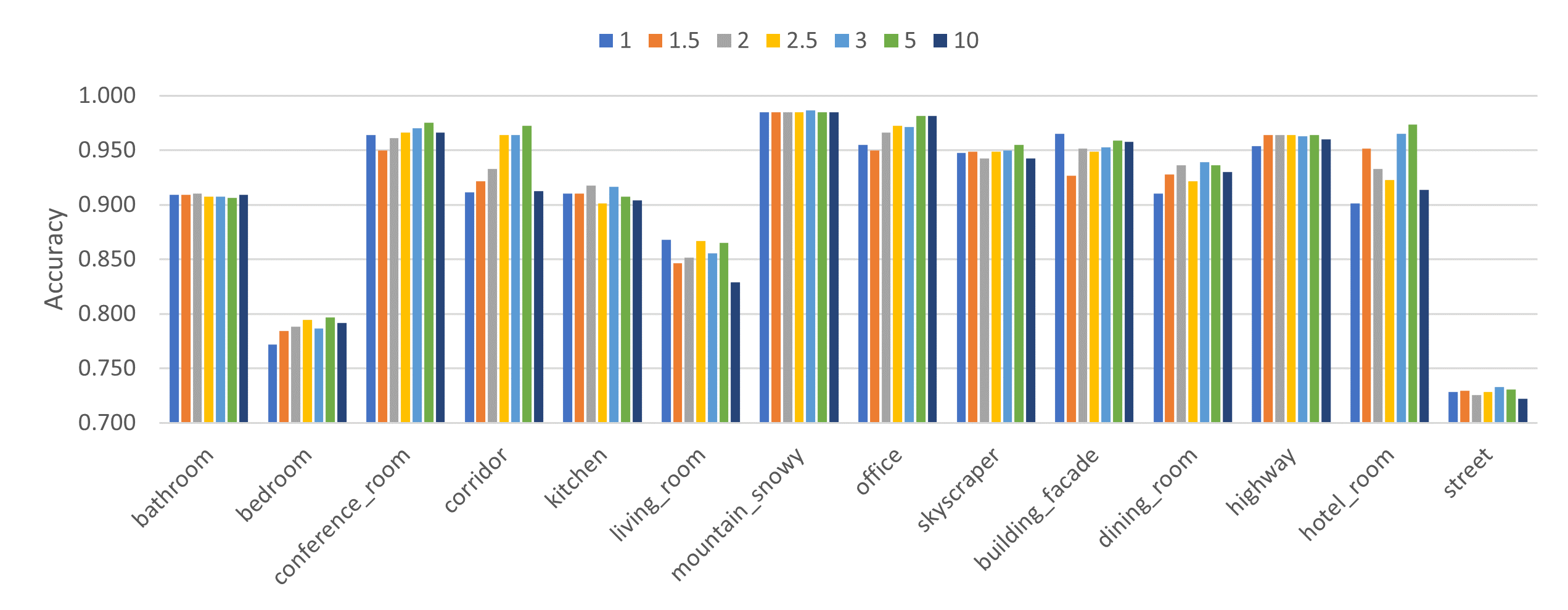}
    \caption{\textbf{Up-weighting accuracy on the remaining classes in ADE20K when using CLIP.} This graph depicts the accuracies when using a CLIP encoder and scaling the loss for low-frequency classes by factors of $1.5, 2, 2.5, 3, 5, $ and $10$. The factor of $1$ corresponds to the baseline accuracy. Similar to the results from the Planet dataset, we find that there is no consistent trend across the scaling factors.}
    \label{fig:clip_ade_loss_nr}
\end{figure}

For single-label classification, the performance on an up-weighted class sometimes drops surprisingly (e.g., with the "game\_room" label in Figure \ref{fig:loss_results_ade20k}).
However, a plausible explanation is that the training examples for that label do not reflect the testing distribution, and the model learns meaningless artifacts and/or makes a faulty association \cite{artifacts1,artifacts2,artifacts3,artifacts4}.
For example, if the training examples for the class "game\_room" tend to contain pool tables and if the testing examples for "game\_room" do not contain pool tables, the model could develop a dependence on the presence of pool tables to identify a scene as "game\_room" and consequently misclassify the testing examples.
This hypothesis could be investigated further by comparing the saliency maps generated for the training images to those generated for the testing images \cite{saliency1,saliency2,saliency3}.

In addition to our experiments, this finding is partly supported by prior experimentation in Byrd \& Lipton's work on importance sampling \cite{byrd2019}.
Byrd \& Lipton use a reweighting scheme that weights each class according to its relative frequency in the dataset.
This scheme is slightly different from our setup in that we use a constant scaling factor for all classes, regardless of frequency.
Nonetheless, Byrd \& Lipton found that the impact of reweighting the loss diminished over time, and our results are consistent with this finding. 




\subsection{Undersampling produces a comparable performance to the baseline on the low-frequency classes while reducing the performance on the other classes.}

\begin{figure}[h!]
\centering
    \begin{subfigure}{\linewidth}
         \centering
         \includegraphics[width=\linewidth]{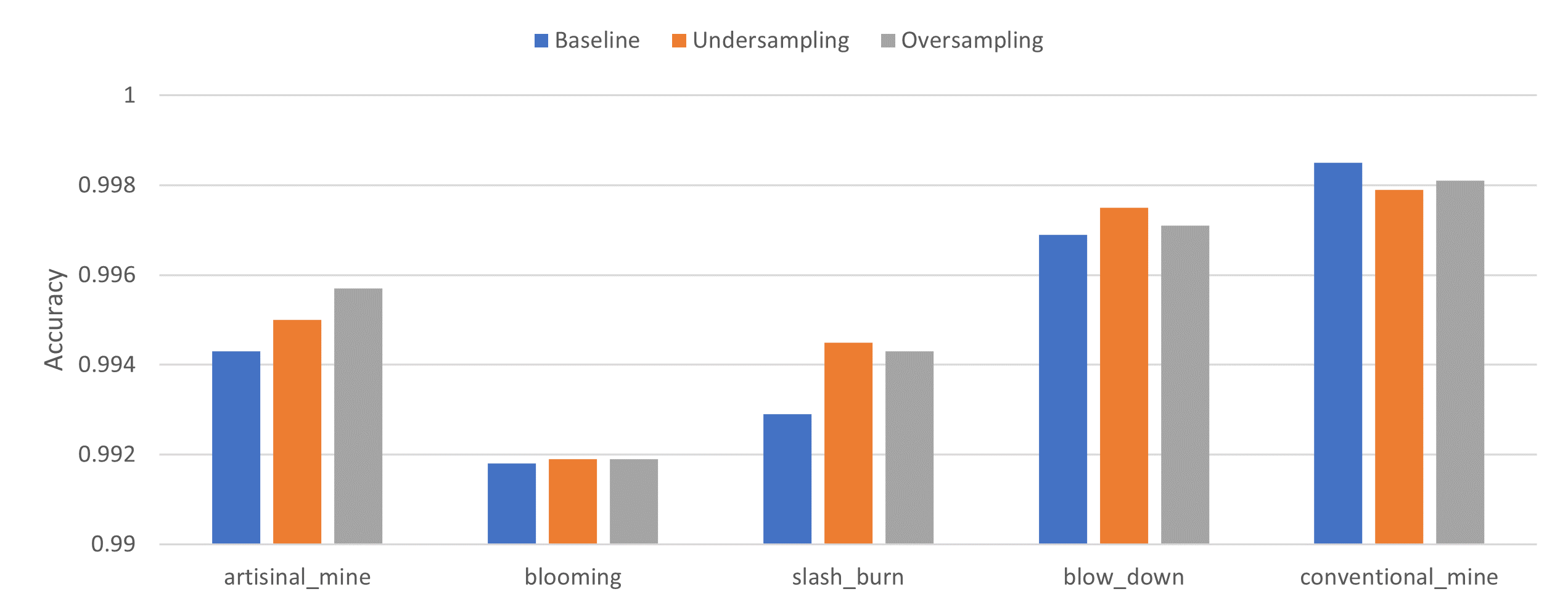}
         \caption{\textbf{Resampling accuracy on Planet's low-frequency classes with CLIP encoder.}}
         \label{fig:clip_for_samp}
    \end{subfigure}

    \vspace{\baselineskip}
    \vfill

    \begin{subfigure}{\linewidth}
         \centering
         \includegraphics[width=\linewidth]{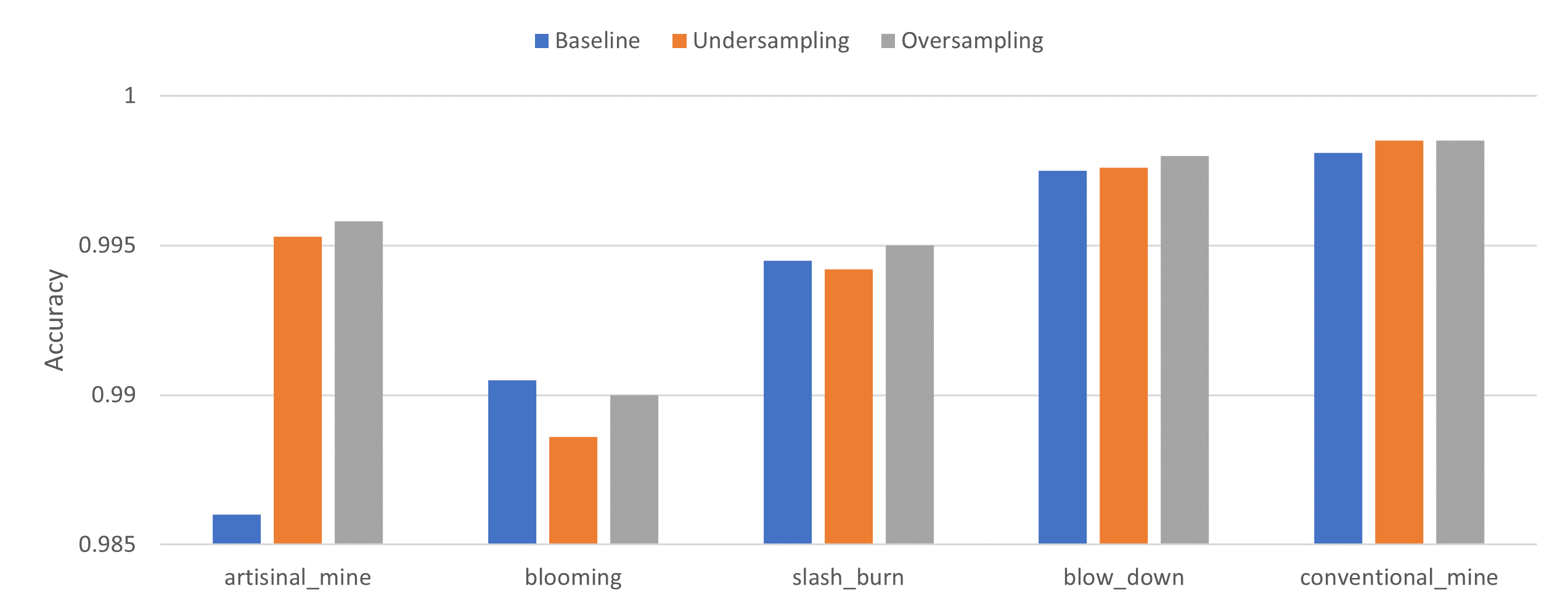}
         \caption{\textbf{Resampling accuracy on Planet's low-frequency classes with ResNet-18 encoder.}}
         \label{fig:resnet_for_samp}
    \end{subfigure}
    \caption{\textbf{Resampling accuracy on the low frequency classes in the Planet dataset.} The classes are sorted in descending order by frequency along the x-axis.}
    \label{fig:resample_results_planet}
\end{figure}

Across both models, the performance when using undersampling roughly matches or surpasses the baseline performance on low-frequency classes (Figure \ref{fig:resample_results_planet}). 
The situations where the performance improves compared to the baseline performance are surprising because we use the same images for these classes.
One potential cause is stochasticity of training which could be resolved by instead training multiple undersampling and baseline models and averaging the performances assuming access to sufficient computational and storage resources.

\begin{figure}[h!]
\centering
    \begin{subfigure}{\linewidth}
         \centering
         \includegraphics[width=\linewidth]{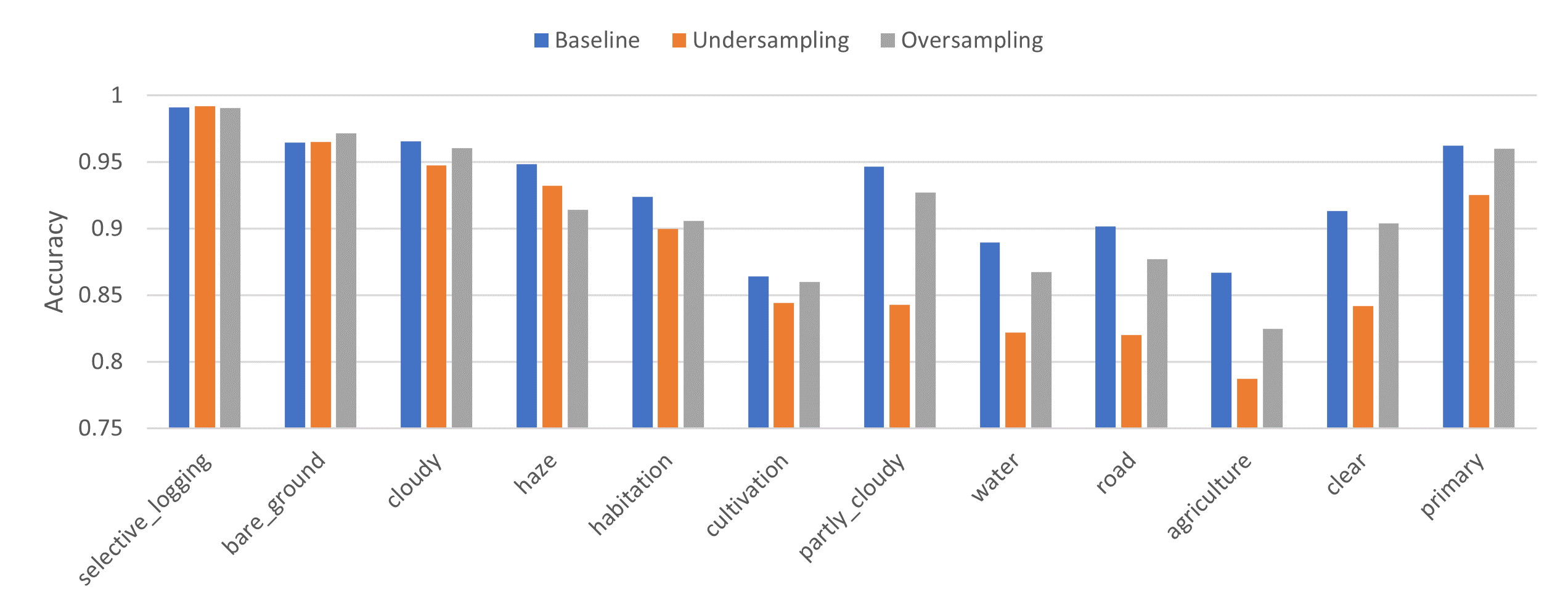}
         \caption{\textbf{Resampling accuracy on Planet's remaining classes with CLIP encoder.}}
         \label{fig:clip_for_samp_nr}
    \end{subfigure}

    \vspace{\baselineskip}
    \vfill

    \begin{subfigure}{\linewidth}
         \centering
         \includegraphics[width=\linewidth]{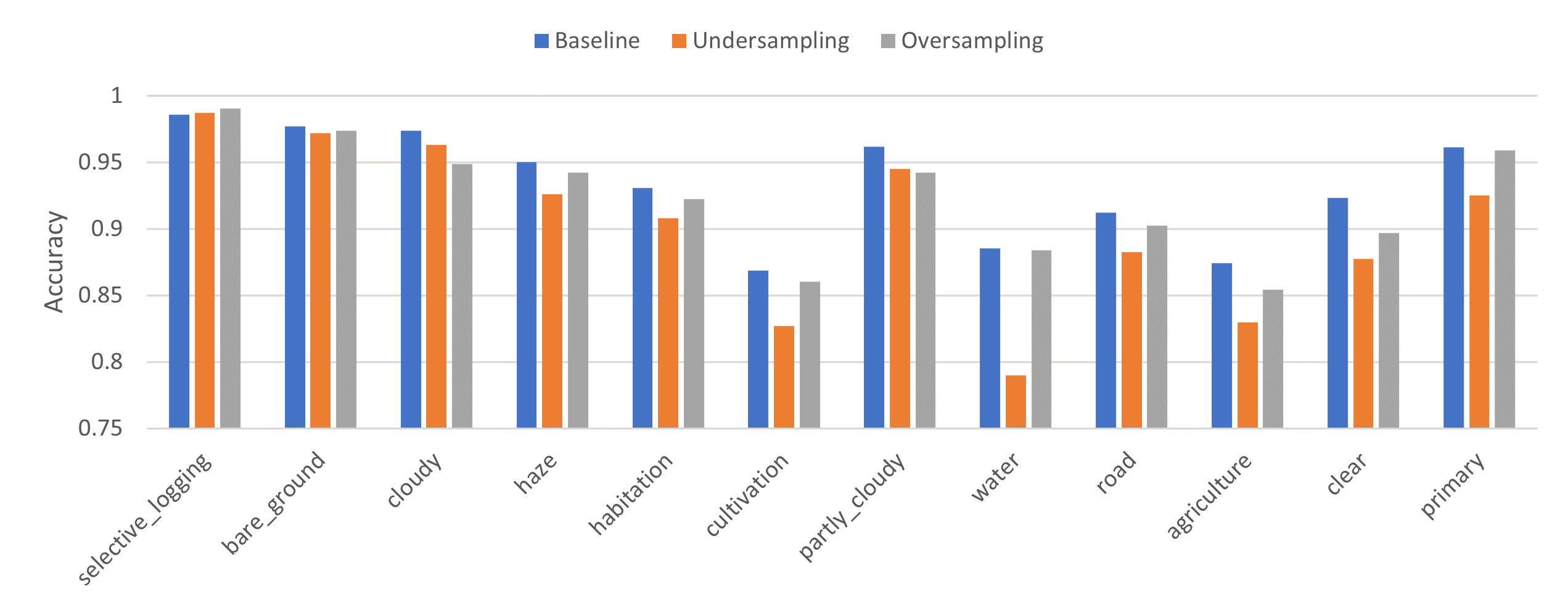}
         \caption{\textbf{Resampling accuracy on Planet's remaining classes with ResNet-18 encoder.}}
         \label{fig:resnet_for_samp_nr}
    \end{subfigure}
    \caption{\textbf{Resampling accuracy on the remaining classes in the Planet dataset.} The classes are sorted in ascending order by frequency along the x-axis.}
    \label{fig:resample_results_planet_nr}
\end{figure}

The performance on the remaining classes often drops (e.g., as with the "agriculture" and "partly\_cloudy" labels) (Figure \ref{fig:resample_results_planet_nr}).
This trend seems reasonable since undersampling removes data from these classes.
However, it is interesting to note that the performance on some of these classes remains roughly the same (e.g., the "selective\_logging" and "bare\_ground" labels).
This trend suggests that the examples removed from those classes were redundant with the examples that remained and that the examples that were removed from the negatively-impacted classes did not contain redundant information \cite{dataredundancy1, dataredundancy2, dataredundancy3, dataredundancy4}.
Furthermore, we use approximately $21$ times less data when undersampling, so the fact that the performance is still reasonable across all classes is impressive and further supports the idea that there may exist some redundancy in the data.



\subsection{Oversampling generally improves performance across low-frequency classes.}

Among the low-frequency classes, oversampling generally matches or improves upon the baseline performance (Figure \ref{fig:resample_results_planet}).
Since oversampling artificially adds data by sampling from existing data, marginal improvements are reasonable to expect.
For situations where the performance is matched (e.g., "blooming" and "blow\_down" in Figure \ref{fig:resample_results_planet}), the repetition of the data could have outweighed any improvement.
In particular, it is interesting to note that, with ResNet-18, as the classes reduce in frequency (progress along the x-axis), the gap between the baseline and oversampling performance reduces (Figure \ref{fig:resnet_for_samp}).
This trend likely occurs because the fewer examples a class contains, the more frequently those examples will be seen during training with oversampling, so the model learns highly repetitive information about that class.

Surprisingly, oversampling sometimes leads to a drop in performance in the remaining classes, though to a lesser degree than undersampling (Figure \ref{fig:resample_results_planet_nr}).
This trend likely occurs because we stopped training prematurely.
Oversampling uses approximately $8$ times more data than the baseline, so using the same number of epochs when training would not be sufficient.
However, as noted in Section \ref{sec:experiments}, we train all models under the same hyperparameters to control for comparison.

\subsection{Comparison between methods}
Surprisingly, up-weighting the loss on low-frequency classes, even by a relatively large factor of $10$, does not produce consistent effects on the performance of any of the classes.
In fact, the two resampling methods seem to produce more consistent effects than reweighting.
Compared to undersampling, oversampling appears to have the potential to cause the most improvement provided more training time to account for the significant increase in data.
However, the performance when undersampling is surprisingly high considering that we use $8$ times less data from Planet and perhaps selecting data according to some criteria rather than at random would help.



\section{Conclusion}
\label{sec:conclusion}

In this work, we present a comparison of methods for mitigating the effect of class imbalance that are based on the idea of importance sampling.
Specifically, we focus on a satellite imagery dataset and provide supplemental experimentation with a scene classification dataset.
Across two types of encoders, CLIP and ResNet-18, we find that up-weighting the loss on low-frequency classes has a negligible effect on the performance.
Additionally, we find that oversampling appears to be more effective than undersampling.
However, we also find that undersampling could still be a valuable method though perhaps users should not select data at random.


In our discussion, we often noted that there likely exists some redundancy in the data.
This idea is part of the motivation for data distillation which attempts to reduce a dataset to a few representative examples per class \cite{distillation,distillation2,distillation3}.
Thus, this could be an interesting alternative to undersampling via random selection.

Additionally, we focus on bias due to underrepresentation.
However, in our discussion, we also hint about bias due to spurious correlation (i.e., when the model forms a faulty association between certain image features and a class) \cite{wang2023pretrainbias}.
Thus, there exist other forms of bias to consider to develop truly robust models, and as machine learning research heads towards a direction of minimizing bias, training data plays an ever more important role in shaping the representations formed through learned parameters.

Future work might also explore whether model accuracy improves sharply for a label once that label's training data size crosses above a certain threshold, suggesting the existence of activation-like behavior at that cut-off. Alternatively, the model accuracy might not become asymptotic --- except as it approaches 100\% --- in the quantity of data per label and instead accuracy could increase continually. A study of this behavior across various leading models could provide useful insights on the amount of data per class needed to achieve certain performance expectations. 

A macroscopic implication of these results is that, as a field, our understanding of how black-box deep neural networks operate is still nascent.
As shown in this report, some of our intuitions on what should improve model performance (e.g., ``more data should improve model performance'' and ``up-weighting loss should improve performance'') do not necessarily hold. 
As machine learning rapidly expands to other fields, the importance of data efficiency and model interpretability grows increasingly important both for generalizability and for trusting AI systems.

\paragraph{Limitations.} In our experiments with up-weighting the loss, we scale the loss for each low-frequency class by the same constant factor. Another option is to scale each class by a factor that is proportional to the relative frequency of the class. Then, among the low-frequency classes, the lowest-frequency class would be up-weighted more than the more-frequent classes. However, Byrd 
\& Lipton still find that this does not improve performance \cite{byrd2019}.

Our model architectures used CLIP and ResNet-18 encoders pretrained on ImageNet \cite{imagenet}.
Satellite imagery comes from a starkly different domain than ImageNet, so one potential improvement would be to train the encoders.
However, previous work has shown that pretraining is still beneficial, and we did use the ADE20K dataset which comes from a similar domain as ImageNet.




\pagebreak

{
    \small
    \bibliographystyle{ieeenat_fullname}
    \bibliography{main}

\begin{thebibliography}{63}
\providecommand{\natexlab}[1]{#1}
\providecommand{\url}[1]{\texttt{#1}}
\expandafter\ifx\csname urlstyle\endcsname\relax
  \providecommand{\doi}[1]{doi: #1}\else
  \providecommand{\doi}{doi: \begingroup \urlstyle{rm}\Url}\fi

\bibitem[Birodkar et~al.(2019)Birodkar, Mobahi, and Bengio]{dataredundancy3}
Vighnesh Birodkar, Hossein Mobahi, and Samy Bengio.
\newblock Semantic redundancies in image-classification datasets: The 10{\%} you don't need.
\newblock \emph{CoRR}, abs/1901.11409, 2019.

\bibitem[Brenskelle et~al.(2020)Brenskelle, Guralnick, Denslow, and Stucky]{botany}
Laura Brenskelle, Rob~P Guralnick, Michael Denslow, and Brian~J Stucky.
\newblock {Maximizing human effort for analyzing scientific images: A case study using digitized herbarium sheets}.
\newblock \emph{Applications in plant sciences}, 8\penalty0 (6):\penalty0 e11370--e11370, 2020.

\bibitem[Buda et~al.(2018)Buda, Maki, and Mazurowski]{classimbalance1}
Mateusz Buda, Atsuto Maki, and Maciej~A. Mazurowski.
\newblock {A systematic study of the class imbalance problem in convolutional neural networks}.
\newblock \emph{Neural Networks}, 106:\penalty0 249--259, 2018.

\bibitem[Buolamwini and Gebru(2018)]{bias1}
Joy Buolamwini and Timnit Gebru.
\newblock {Gender Shades: Intersectional Accuracy Disparities in Commercial Gender Classification}.
\newblock In \emph{Proceedings of the 1st Conference on Fairness, Accountability and Transparency}, pages 77--91. PMLR, 2018.

\bibitem[Byrd and Lipton(2019)]{byrd2019}
Jonathon Byrd and Zachary~Chase Lipton.
\newblock {What is the Effect of Importance Weighting in Deep Learning?}
\newblock In \emph{{Proceedings of the 36th International Conference on Machine Learning}}, pages 872--881. {PMLR}, 2019.

\bibitem[Cazenavette et~al.(2022)Cazenavette, Wang, Torralba, Efros, and Zhu]{distillation2}
George Cazenavette, Tongzhou Wang, Antonio Torralba, Alexei~A. Efros, and Jun-Yan Zhu.
\newblock {Dataset Distillation by Matching Training Trajectories}.
\newblock In \emph{Proceedings of the IEEE/CVF Conference on Computer Vision and Pattern Recognition (CVPR)}, pages 10718--10727, 2022.

\bibitem[Creswell et~al.(2018)Creswell, White, Dumoulin, Arulkumaran, Sengupta, and Bharath]{gans1}
Antonia Creswell, Tom White, Vincent Dumoulin, Kai Arulkumaran, Biswa Sengupta, and Anil~A. Bharath.
\newblock {Generative Adversarial Networks: An Overview}.
\newblock \emph{IEEE Signal Processing Magazine}, 35\penalty0 (1):\penalty0 53--65, 2018.

\bibitem[Cui et~al.(2019)Cui, Jia, Lin, Song, and Belongie]{cui_class-balanced_2019}
Yin Cui, Menglin Jia, Tsung-Yi Lin, Yang Song, and Serge Belongie.
\newblock Class-{Balanced} {Loss} {Based} on {Effective} {Number} of {Samples}, 2019.
\newblock arXiv:1901.05555 [cs].

\bibitem[Fernández et~al.(2018)Fernández, García, Galar, Prati, Krawczyk, and Herrera]{fernandez_cost-sensitive_2018}
Alberto Fernández, Salvador García, Mikel Galar, Ronaldo~C. Prati, Bartosz Krawczyk, and Francisco Herrera.
\newblock Cost-{Sensitive} {Learning}.
\newblock In \emph{Learning from {Imbalanced} {Data} {Sets}}, pages 63--78. Springer International Publishing, Cham, 2018.

\bibitem[Fong and Vedaldi(2017)]{artifacts4}
R. Fong and A. Vedaldi.
\newblock {Interpretable Explanations of Black Boxes by Meaningful Perturbation}.
\newblock In \emph{International Conference on Computer Vision (ICCV)}, 2017.

\bibitem[Goldenberg et~al.(2017)Goldenberg, Uzkent, Clough, Funke, Desai, grischa, Martinez~Manso, Scott, Risdal, Ryan, Pete, Holm, Nair, Herron, Stafford, and Kan]{labs_planet_2017}
Benjamin Goldenberg, Burak Uzkent, Christian Clough, Dennis Funke, Deven Desai, grischa, Jesus Martinez~Manso, Kat Scott, Meg Risdal, Mike Ryan, Pete, Rachel Holm, Ramesh Nair, Sean Herron, Tony Stafford, and Wendy Kan.
\newblock {Planet: Understanding the Amazon from Space}, 2017.

\bibitem[Gong and Kim(2017)]{gong_rhsboost_2017}
Joonho Gong and Hyunjoong Kim.
\newblock {RHSBoost}: {Improving} classification performance in imbalance data.
\newblock \emph{Computational Statistics \& Data Analysis}, 111:\penalty0 1--13, 2017.

\bibitem[Goodfellow et~al.(2014)Goodfellow, Pouget-Abadie, Mirza, Xu, Warde-Farley, Ozair, Courville, and Bengio]{gans2}
Ian Goodfellow, Jean Pouget-Abadie, Mehdi Mirza, Bing Xu, David Warde-Farley, Sherjil Ozair, Aaron Courville, and Yoshua Bengio.
\newblock {Generative Adversarial Nets}.
\newblock In \emph{Advances in Neural Information Processing Systems}. Curran Associates, Inc., 2014.

\bibitem[Gupta et~al.(2019)Gupta, Dollar, and Girshick]{lvis}
Agrim Gupta, Piotr Dollar, and Ross Girshick.
\newblock {LVIS: A Dataset for Large Vocabulary Instance Segmentation}.
\newblock In \emph{Proceedings of the IEEE/CVF Conference on Computer Vision and Pattern Recognition (CVPR)}, 2019.

\bibitem[{He} et~al.(2016){He}, {Zhang}, {Ren}, and {Sun}]{resnet}
K. {He}, X. {Zhang}, S. {Ren}, and J. {Sun}.
\newblock Deep residual learning for image recognition.
\newblock In \emph{2016 IEEE Conference on Computer Vision and Pattern Recognition (CVPR)}, pages 770--778, 2016.

\bibitem[Herland et~al.(2018)Herland, Khoshgoftaar, and Bauder]{herland_big_2018}
Matthew Herland, Taghi~M. Khoshgoftaar, and Richard~A. Bauder.
\newblock Big {Data} fraud detection using multiple medicare data sources.
\newblock \emph{Journal of Big Data}, 5\penalty0 (1):\penalty0 29, 2018.

\bibitem[Horn et~al.(2018)Horn, Aodha, Song, Cui, Sun, Shepard, Adam, Perona, and Belongie]{inaturalist}
G.~Van Horn, O.~Mac Aodha, Y. Song, Y. Cui, C. Sun, A. Shepard, H. Adam, P. Perona, and S. Belongie.
\newblock {The iNaturalist Species Classification and Detection Dataset}.
\newblock In \emph{2018 IEEE/CVF Conference on Computer Vision and Pattern Recognition (CVPR)}, pages 8769--8778, Los Alamitos, CA, USA, 2018. IEEE Computer Society.

\bibitem[Horvitz and Thompson(1952)]{importance1}
D.~G. Horvitz and D.~J. Thompson.
\newblock A generalization of sampling without replacement from a finite universe.
\newblock \emph{Journal of the American Statistical Association}, 47\penalty0 (260):\penalty0 663--685, 1952.

\bibitem[Huang et~al.(2016)Huang, Li, Loy, and Tang]{classimbalance4}
Chen Huang, Yining Li, Chen~Change Loy, and Xiaoou Tang.
\newblock Learning deep representation for imbalanced classification.
\newblock In \emph{Proceedings of the IEEE Conference on Computer Vision and Pattern Recognition (CVPR)}, 2016.

\bibitem[Jaiswal et~al.(2021)Jaiswal, Babu, Zadeh, Banerjee, and Makedon]{selfsuplearning2}
Ashish Jaiswal, Ashwin~Ramesh Babu, Mohammad~Zaki Zadeh, Debapriya Banerjee, and Fillia Makedon.
\newblock {A Survey on Contrastive Self-Supervised Learning}.
\newblock \emph{Technologies}, 9\penalty0 (1), 2021.

\bibitem[Jiang et~al.(2022)Jiang, Najibi, Qi, Zhou, and Anguelov]{choosingrare}
Chiyu~Max Jiang, Mahyar Najibi, Charles~R. Qi, Yin Zhou, and Dragomir Anguelov.
\newblock Improving the intra-class long-tail in 3d detection via rare example mining.
\newblock In \emph{Computer Vision -- ECCV 2022}, pages 158--175, Cham, 2022. Springer Nature Switzerland.

\bibitem[Johnson and Khoshgoftaar(2019)]{classimbalance2}
Justin~M. Johnson and Taghi~M. Khoshgoftaar.
\newblock {Survey on deep learning with class imbalance}.
\newblock \emph{Journal of Big Data}, 6\penalty0 (1):\penalty0 27, 2019.

\bibitem[Kahn and Marshall(1953)]{importance2}
H. Kahn and A.~W. Marshall.
\newblock Methods of reducing sample size in monte carlo computations.
\newblock \emph{Journal of the Operations Research Society of America}, 1\penalty0 (5):\penalty0 263--278, 1953.

\bibitem[Kaushal et~al.(2019)Kaushal, Iyer, Kothawade, Mahadev, Doctor, and Ramakrishnan]{dataredundancy4}
Vishal Kaushal, Rishabh Iyer, Suraj Kothawade, Rohan Mahadev, Khoshrav Doctor, and Ganesh Ramakrishnan.
\newblock {Learning From Less Data: A Unified Data Subset Selection and Active Learning Framework for Computer Vision}.
\newblock In \emph{2019 IEEE Winter Conference on Applications of Computer Vision (WACV)}, pages 1289--1299, 2019.

\bibitem[Khosla et~al.(2012)Khosla, Zhou, Malisiewicz, Efros, and Torralba]{bias2}
Aditya Khosla, Tinghui Zhou, Tomasz Malisiewicz, Alexei~A. Efros, and Antonio Torralba.
\newblock Undoing the damage of dataset bias.
\newblock In \emph{Computer Vision -- ECCV 2012}, pages 158--171, Berlin, Heidelberg, 2012. Springer Berlin Heidelberg.

\bibitem[Koller and Friedman(2009)]{importance4}
Daphne Koller and Nir Friedman.
\newblock \emph{Probabilistic Graphical Models: Principles and Techniques - Adaptive Computation and Machine Learning}.
\newblock The MIT Press, 2009.

\bibitem[Kubat(2000)]{kubat_addressing_2000}
M. Kubat.
\newblock Addressing the {Curse} of {Imbalanced} {Training} {Sets}: {One}-{Sided} {Selection}.
\newblock \emph{Fourteenth International Conference on Machine Learning}, 2000.

\bibitem[Kubat et~al.(1998)Kubat, Holte, and Matwin]{kubat_machine_1998}
Miroslav Kubat, Robert~C. Holte, and Stan Matwin.
\newblock Machine {Learning} for the {Detection} of {Oil} {Spills} in {Satellite} {Radar} {Images}.
\newblock \emph{Machine Learning}, 30\penalty0 (2):\penalty0 195--215, 1998.

\bibitem[Kurakin et~al.(2017)Kurakin, Goodfellow, and Bengio]{artifacts2}
Alexey Kurakin, Ian Goodfellow, and Samy Bengio.
\newblock Adversarial examples in the physical world, 2017.

\bibitem[Lee(2013)]{lee2013}
Dong-Hyun Lee.
\newblock Pseudo-label : The simple and efficient semi-supervised learning method for deep neural networks.
\newblock \emph{ICML 2013 Workshop : Challenges in Representation Learning (WREPL)}, 2013.

\bibitem[Liu et~al.(2009)Liu, Wu, and Zhou]{undersamplethreshold}
Xu-Ying Liu, Jianxin Wu, and Zhi-Hua Zhou.
\newblock {Exploratory Undersampling for Class-Imbalance Learning}.
\newblock \emph{IEEE Transactions on Systems, Man, and Cybernetics, Part B (Cybernetics)}, 39\penalty0 (2):\penalty0 539--550, 2009.

\bibitem[Liu et~al.(2019)Liu, Miao, Zhan, Wang, Gong, and Yu]{longtail2}
Ziwei Liu, Zhongqi Miao, Xiaohang Zhan, Jiayun Wang, Boqing Gong, and Stella~X. Yu.
\newblock {Large-Scale Long-Tailed Recognition in an Open World}.
\newblock In \emph{IEEE Conference on Computer Vision and Pattern Recognition (CVPR)}, 2019.

\bibitem[Luque et~al.(2019)Luque, Carrasco, Martín, and de~las Heras]{luque_impact_2019}
Amalia Luque, Alejandro Carrasco, Alejandro Martín, and Ana de~las Heras.
\newblock The impact of class imbalance in classification performance metrics based on the binary confusion matrix.
\newblock \emph{Pattern Recognition}, 91:\penalty0 216--231, 2019.

\bibitem[Mahendran and Vedaldi(2015)]{artifacts3}
Aravindh Mahendran and Andrea Vedaldi.
\newblock {Understanding Deep Image Representations by Inverting Them}.
\newblock In \emph{IEEE Conference on Computer Vision and Pattern Recognition (CVPR)}, 2015.

\bibitem[Nguyen et~al.(2015)Nguyen, Yosinski, and Clune]{artifacts1}
Anh Nguyen, Jason Yosinski, and Jeff Clune.
\newblock {Deep Neural Networks are Easily Fooled: High Confidence Predictions for Unrecognizable Images}.
\newblock In \emph{IEEE Conference on Computer Vision and Pattern Recognition (CVPR)}, 2015.

\bibitem[Nguyen et~al.(2021)Nguyen, Novak, Xiao, and Lee]{dataredundancy2}
Timothy Nguyen, Roman Novak, Lechao Xiao, and Jaehoon Lee.
\newblock {Dataset Distillation with Infinitely Wide Convolutional Networks}.
\newblock In \emph{Advances in Neural Information Processing Systems}, pages 5186--5198. Curran Associates, Inc., 2021.

\bibitem[Parisot et~al.(2022)Parisot, Esperan\c{c}a, McDonagh, Madarasz, Yang, and Li]{choosingrare2}
Sarah Parisot, Pedro~M. Esperan\c{c}a, Steven McDonagh, Tamas~J. Madarasz, Yongxin Yang, and Zhenguo Li.
\newblock {Long-Tail Recognition via Compositional Knowledge Transfer}.
\newblock In \emph{Proceedings of the IEEE/CVF Conference on Computer Vision and Pattern Recognition (CVPR)}, pages 6939--6948, 2022.

\bibitem[Prati et~al.(2015)Prati, Batista, and Silva]{prati_class_2015}
Ronaldo~C. Prati, Gustavo E. A. P.~A. Batista, and Diego~F. Silva.
\newblock Class imbalance revisited: a new experimental setup to assess the performance of treatment methods.
\newblock \emph{Knowledge and Information Systems}, 45\penalty0 (1):\penalty0 247--270, 2015.

\bibitem[Radford et~al.(2021)Radford, Kim, Hallacy, Ramesh, Goh, Agarwal, Sastry, Askell, Mishkin, Clark, Krueger, and Sutskever]{clip}
Alec Radford, Jong~Wook Kim, Chris Hallacy, Aditya Ramesh, Gabriel Goh, Sandhini Agarwal, Girish Sastry, Amanda Askell, Pamela Mishkin, Jack Clark, Gretchen Krueger, and Ilya Sutskever.
\newblock {Learning Transferable Visual Models From Natural Language Supervision}.
\newblock In \emph{Proceedings of the 38th International Conference on Machine Learning}, pages 8748--8763. PMLR, 2021.

\bibitem[Radosavovic et~al.(2018)Radosavovic, Dollár, Girshick, Gkioxari, and He]{distillation}
Ilija Radosavovic, Piotr Dollár, Ross Girshick, Georgia Gkioxari, and Kaiming He.
\newblock {Data Distillation: Towards Omni-Supervised Learning}.
\newblock In \emph{Proceedings of the IEEE Conference on Computer Vision and Pattern Recognition (CVPR)}, 2018.

\bibitem[Rao et~al.(2006)Rao, Krishnan, and Niculescu]{rao_data_2006}
R.~Bharat Rao, Sriram Krishnan, and Radu~Stefan Niculescu.
\newblock Data mining for improved cardiac care.
\newblock \emph{ACM SIGKDD Explorations Newsletter}, 8\penalty0 (1):\penalty0 3--10, 2006.

\bibitem[Rom(2019)]{rom_planets_dataset_2019}
Nikita Rom.
\newblock planets\_dataset, 2019.

\bibitem[Rubinstein and Kroese(2016)]{importance3}
Reuven~Y. Rubinstein and Dirk~P. Kroese.
\newblock \emph{Simulation and the Monte Carlo Method}.
\newblock Wiley Publishing, 3rd edition, 2016.

\bibitem[Russakovsky et~al.(2015)Russakovsky, Deng, Su, Krause, Satheesh, Ma, Huang, Karpathy, Khosla, Bernstein, Berg, and Fei-Fei]{imagenet}
Olga Russakovsky, Jia Deng, Hao Su, Jonathan Krause, Sanjeev Satheesh, Sean Ma, Zhiheng Huang, Andrej Karpathy, Aditya Khosla, Michael Bernstein, Alexander~C Berg, and Li Fei-Fei.
\newblock {ImageNet Large Scale Visual Recognition Challenge}.
\newblock 2015.

\bibitem[Sachdeva and McAuley(2023)]{distillation3}
Noveen Sachdeva and Julian McAuley.
\newblock {Data Distillation: A Survey}.
\newblock \emph{Transactions on Machine Learning Research}, 2023.
\newblock Survey Certification.

\bibitem[Sampath et~al.(2021)Sampath, Maurtua, Aguilar~Mart{\'\i}n, and Gutierrez]{classimbalance3}
Vignesh Sampath, I{\~n}aki Maurtua, Juan~Jos{\'e} Aguilar~Mart{\'\i}n, and Aitor Gutierrez.
\newblock A survey on generative adversarial networks for imbalance problems in computer vision tasks.
\newblock \emph{Journal of Big Data}, 8\penalty0 (1):\penalty0 27, 2021.

\bibitem[Santos et~al.(2020)Santos, Ferreira, and Silva]{geosciences}
Jéssica~S. Santos, Rodrigo~S. Ferreira, and Viviane~T. Silva.
\newblock Evaluating the classification of images from geoscience papers using small data.
\newblock \emph{Applied Computing and Geosciences}, 5:\penalty0 100018, 2020.

\bibitem[Selvaraju et~al.(2017)Selvaraju, Cogswell, Das, Vedantam, Parikh, and Batra]{saliency1}
Ramprasaath~R. Selvaraju, Michael Cogswell, Abhishek Das, Ramakrishna Vedantam, Devi Parikh, and Dhruv Batra.
\newblock {{Grad-CAM}: Visual Explanations From Deep Networks via Gradient-Based Localization}.
\newblock In \emph{International Conference on Computer Vision (ICCV)}, 2017.

\bibitem[Shitole et~al.(2021)Shitole, Li, Kahng, Tadepalli, and Fern]{saliency3}
Vivswan Shitole, Fuxin Li, Minsuk Kahng, Prasad Tadepalli, and Alan Fern.
\newblock {One Explanation is Not Enough: Structured Attention Graphs for Image Classification}.
\newblock In \emph{Neural Information Processing Systems (NeurIPS)}, 2021.

\bibitem[Shu et~al.(2019)Shu, Xie, Yi, Zhao, Zhou, Xu, and Meng]{shu_meta-weight-net_2019}
Jun Shu, Qi Xie, Lixuan Yi, Qian Zhao, Sanping Zhou, Zongben Xu, and Deyu Meng.
\newblock Meta-{Weight}-{Net}: {Learning} an {Explicit} {Mapping} {For} {Sample} {Weighting}, 2019.
\newblock arXiv:1902.07379 [cs, stat].

\bibitem[Simonyan et~al.(2014)Simonyan, Vedaldi, and Zisserman]{saliency2}
Karen Simonyan, Andrea Vedaldi, and Andrew Zisserman.
\newblock Deep inside convolutional networks: Visualising image classification models and saliency maps.
\newblock In \emph{International Conference on Learning Representations (ICLR) Workshops}, 2014.

\bibitem[Thai-Nghe et~al.(2009)Thai-Nghe, Busche, and Schmidt-Thieme]{thai-nghe_improving_2009}
Nguyen Thai-Nghe, Andre Busche, and Lars Schmidt-Thieme.
\newblock Improving {Academic} {Performance} {Prediction} by {Dealing} with {Class} {Imbalance}.
\newblock In \emph{2009 {Ninth} {International} {Conference} on {Intelligent} {Systems} {Design} and {Applications}}, pages 878--883, 2009.
\newblock ISSN: 2164-7151.

\bibitem[Tommasi et~al.(2017)Tommasi, Patricia, Caputo, and Tuytelaars]{bias3}
Tatiana Tommasi, Novi Patricia, Barbara Caputo, and Tinne Tuytelaars.
\newblock \emph{A Deeper Look at Dataset Bias}, pages 37--55.
\newblock Springer International Publishing, Cham, 2017.

\bibitem[Wang and Russakovsky(2023)]{wang2023pretrainbias}
Angelina Wang and Olga Russakovsky.
\newblock {Overcoming Bias in Pretrained Models by Manipulating the Finetuning Dataset}.
\newblock 2023.

\bibitem[Wang et~al.(2021)Wang, Li, Wang, Liu, Wang, Tan, Wu, Liu, Sun, Yang, Liu, Chen, Zhou, {Ben Ayed}, and Zheng]{medicine}
Shanshan Wang, Cheng Li, Rongpin Wang, Zaiyi Liu, Meiyun Wang, Hongna Tan, Yaping Wu, Xinfeng Liu, Hui Sun, Rui Yang, Xin Liu, Jie Chen, Huihui Zhou, Ismail {Ben Ayed}, and Hairong Zheng.
\newblock {Annotation-efficient deep learning for automatic medical image segmentation}.
\newblock \emph{Nature Communications}, 12\penalty0 (1):\penalty0 5915, 2021.

\bibitem[Wang et~al.(2020)Wang, Zhu, Torralba, and Efros]{dataredundancy1}
Tongzhou Wang, Jun-Yan Zhu, Antonio Torralba, and Alexei~A. Efros.
\newblock Dataset distillation, 2020.

\bibitem[Wei et~al.(2013)Wei, Li, Cao, Ou, and Chen]{wei_effective_2013}
Wei Wei, Jinjiu Li, Longbing Cao, Yuming Ou, and Jiahang Chen.
\newblock Effective detection of sophisticated online banking fraud on extremely imbalanced data.
\newblock \emph{World Wide Web}, 16\penalty0 (4):\penalty0 449--475, 2013.

\bibitem[Yang et~al.(2023)Yang, Song, King, and Xu]{semisuplearning}
Xiangli Yang, Zixing Song, Irwin King, and Zenglin Xu.
\newblock A survey on deep semi-supervised learning.
\newblock \emph{IEEE Transactions on Knowledge and Data Engineering}, 35\penalty0 (9):\penalty0 8934--8954, 2023.

\bibitem[Zadrozny et~al.(2003)Zadrozny, Langford, and Abe]{zadrozny_cost-sensitive_2003}
B. Zadrozny, J. Langford, and N. Abe.
\newblock Cost-sensitive learning by cost-proportionate example weighting.
\newblock In \emph{Third {IEEE} {International} {Conference} on {Data} {Mining}}, pages 435--442, 2003.

\bibitem[Zhai et~al.(2019)Zhai, Oliver, Kolesnikov, and Beyer]{selfsuplearning}
Xiaohua Zhai, Avital Oliver, Alexander Kolesnikov, and Lucas Beyer.
\newblock {S4L: Self-Supervised Semi-Supervised Learning}.
\newblock In \emph{Proceedings of the IEEE/CVF International Conference on Computer Vision (ICCV)}, 2019.

\bibitem[Zhou et~al.(2017)Zhou, Zhao, Puig, Fidler, Barriuso, and Torralba]{zhou_scene_2017}
Bolei Zhou, Hang Zhao, Xavier Puig, Sanja Fidler, Adela Barriuso, and Antonio Torralba.
\newblock Scene {Parsing} through {ADE20K} {Dataset}.
\newblock In \emph{2017 {IEEE} {Conference} on {Computer} {Vision} and {Pattern} {Recognition} ({CVPR})}, pages 5122--5130, 2017.
\newblock ISSN: 1063-6919.

\bibitem[Zhou et~al.(2018)Zhou, Lapedriza, Khosla, Oliva, and Torralba]{places365}
Bolei Zhou, Agata Lapedriza, Aditya Khosla, Aude Oliva, and Antonio Torralba.
\newblock {Places: A 10 Million Image Database for Scene Recognition}.
\newblock \emph{IEEE Transactions on Pattern Analysis and Machine Intelligence}, 40\penalty0 (6):\penalty0 1452--1464, 2018.

\bibitem[Zhu et~al.(2014)Zhu, Anguelov, and Ramanan]{longtail3}
Xiangxin Zhu, Dragomir Anguelov, and Deva Ramanan.
\newblock {Capturing Long-Tail Distributions of Object Subcategories}.
\newblock In \emph{2014 IEEE Conference on Computer Vision and Pattern Recognition}, pages 915--922, 2014.

\end{thebibliography}
}


\end{document}